\documentclass[10pt,journal,compsoc]{IEEEtran}
\ifCLASSOPTIONcompsoc
  \usepackage[nocompress]{cite}
\else
  \usepackage{cite}
\fi
\ifCLASSINFOpdf
\else
\fi
\usepackage{amsthm}
\usepackage{amssymb}
\usepackage{amsmath}
\usepackage{multirow}
\usepackage[pagebackref=false,breaklinks=true,colorlinks,bookmarks=false]{hyperref}
\usepackage{graphicx}
\usepackage{comment}
\usepackage{enumitem}
\usepackage{booktabs}

\usepackage{pifont}
\newcommand{\cmark}{\ding{51}}%
\newcommand{\xmark}{\ding{55}}%
\newcommand{\romannum}[1]{\romannumeral #1} 

\usepackage{array}
\newcolumntype{x}[1]{>{\centering\arraybackslash}p{#1}}

\newcommand{\keypoint}[1]{\vspace{0.1cm}\noindent\textbf{#1}\hspace{1em}}
\newcommand{\tableCellHeight}{1.1}

\theoremstyle{definition}

\newcommand{\tabstyle}[1]{
  \setlength{\tabcolsep}{#1}
  \renewcommand{\arraystretch}{\tableCellHeight}
  \centering
}

\begin{document}
%
\title{Domain Generalization: A Survey}
%
%
%
%

\author{
Kaiyang Zhou,
Ziwei Liu,
Yu Qiao,
Tao Xiang,
and Chen Change Loy
\IEEEcompsocitemizethanks{
\IEEEcompsocthanksitem K.~Zhou, Z.~Liu and C.C.~Loy are with the S-Lab, Nanyang Technological University, Singapore. E-mail: \{kaiyang.zhou, ziwei.liu, ccloy\}@ntu.edu.sg.\par
\IEEEcompsocthanksitem Y.~Qiao is with Shenzhen Institute of Advanced Technology, Chinese Academy of Sciences, China, and also with Shanghai AI Lab, Shanghai, China. E-mail: yu.qiao@siat.ac.cn.\par
\IEEEcompsocthanksitem T.~Xiang is with the Centre for Vision Speech and Signal Processing, University of Surrey, Guildford, UK. E-mail: t.xiang@surrey.ac.uk.
}
}

\IEEEtitleabstractindextext{%
\begin{abstract}
Generalization to out-of-distribution (OOD) data is a capability natural to humans yet challenging for machines to reproduce. This is because most learning algorithms strongly rely on the i.i.d.~assumption on source/target data, which is often violated in practice due to domain shift. Domain generalization (DG) aims to achieve OOD generalization by using only source data for model learning. Over the last ten years, research in DG has made great progress, leading to a broad spectrum of methodologies, e.g., those based on domain alignment, meta-learning, data augmentation, or ensemble learning, to name a few; DG has also been studied in various application areas including computer vision, speech recognition, natural language processing, medical imaging, and reinforcement learning. In this paper, for the first time a comprehensive literature review in DG is provided to summarize the developments over the past decade. Specifically, we first cover the background by formally defining DG and relating it to other relevant fields like domain adaptation and transfer learning. Then, we conduct a thorough review into existing methods and theories. Finally, we conclude this survey with insights and discussions on future research directions.
\end{abstract}

\begin{IEEEkeywords}
Out-of-Distribution Generalization, Domain Shift, Model Robustness, Machine Learning
\end{IEEEkeywords}}

\maketitle

\IEEEdisplaynontitleabstractindextext

%
\IEEEpeerreviewmaketitle

\IEEEraisesectionheading{\section{Introduction}\label{sec:introduction}}
\IEEEPARstart{I}{f} an image classifier was trained on photo images, would it work on sketch images? What if a car detector trained using urban images is tested in rural environments? Is it possible to deploy a semantic segmentation model trained using sunny images under rainy or snowy weather conditions? Can a health status classifier trained using one patient's electrocardiogram data be used to diagnose another patient's health status? Answers to all these questions depend on how well the machine learning models can deal with one common problem, namely the \emph{domain shift} problem. Such a problem refers to the distribution shift between  a set of training (source) data and a set of test (target) data~\cite{moreno2012unifying,recht2019imagenet,ben2010theory,taori2020measuring,blanchard2021domain}.

Most statistical learning algorithms strongly rely on an over-simplified assumption, that is, the source and target data are independent and identically distributed (i.i.d.), while ignoring out-of-distribution (OOD) scenarios commonly encountered in practice. This means that they are not designed with the domain shift problem in mind, and as a consequence, a learning agent trained only with source data will typically suffer significant performance drops on an OOD target domain.

The domain shift problem has seriously impeded large-scale deployments of machine learning models. One might be curious if recent advances in deep neural networks~\cite{he2016deep,krizhevsky2012imagenet}, known as deep learning~\cite{lecun2015deep}, can mitigate this problem. Studies in~\cite{recht2019imagenet,hendrycks2019benchmarking,yang2021generalized} suggest that deep learning models' performance degrades significantly on OOD datasets, even with just small variations in the data generating process. This highlights the fact that the successes achieved by deep learning so far have been largely driven by supervised learning with large-scale annotated datasets like ImageNet~\cite{deng2009imagenet}---again, relying on the i.i.d.~assumption.

Research on how to deal with domain shift has been extensively conducted in the literature. A straightforward solution to bypass the OOD data issue is to collect some data from the target domain to adapt a source-domain-trained model. Indeed, this domain adaptation (DA) problem has received much attention~\cite{saenko2010adapting,lu2020stochastic,saito2018maximum,ganin2015unsupervised,long2015learning,liu2020open,li2021learning}. However, DA relies on a strong assumption that target data is accessible for model adaptation, which does not always hold in practice.

In many applications, target data is difficult to obtain or even unknown before deploying the model. For example, in biomedical applications where domain shift occurs between different patients' data, it is impractical to collect each new patient's data in advance~\cite{muandet2013domain}; in traffic scene semantic segmentation it is infeasible to collect data capturing all different scenes and under all possible weather conditions~\cite{yue2019domain}; when dealing with data stream, the model is also required to be intrinsically generalizable~\cite{volpi2021continual}.

To overcome the domain shift problem, as well as the absence of target data, the problem of \emph{domain generalization} (DG) was introduced~\cite{blanchard2011generalizing}. Specifically, the goal in DG is to learn a model using data from a single or multiple related but distinct source domains in such a way that the model can generalize well to any OOD target domain.

Since the first formal introduction in 2011 by Blanchard et al.~\cite{blanchard2011generalizing}, a plethora of methods have been developed to tackle the OOD generalization issue~\cite{zhou2021mixstyle,fan2021adversarially,zhang2021deep,pandey2021domain,shu2021open,zhou2021mixstylenn,zhou2021stylematch,cha2021swad}. This includes methods based on aligning source domain distributions for domain-invariant representation learning~\cite{li2018mmdaae,li2018ciddg}, exposing the model to domain shift during training via meta-learning~\cite{li2018learning,balaji2018metareg}, and augmenting data with domain synthesis~\cite{zhou2020learning,zhou2020deep}, to name a few. From the application point of view, DG has not only been studied in computer vision like object recognition~\cite{li2017deeper,feature_critic}, semantic segmentation~\cite{volpi2019addressing,yue2019domain} and person re-identification~\cite{zhou2020learning,zhou2021mixstyle}, but also in other domains such as speech recognition~\cite{shankar2018generalizing}, natural language processing~\cite{balaji2018metareg}, medical imaging~\cite{liu2020ms,liu2020shape}, and reinforcement learning~\cite{zhou2021mixstyle}.

In this survey, we aim to provide a timely and comprehensive literature review to summarize, mainly from the technical perspective, the learning algorithms developed over the last decade, and provide insights on potential directions for future research.

\section{Background}\label{sec:bg}

\subsection{A Brief History of Domain Generalization}\label{sec:bg;subsec:history_DG}
The domain generalization (DG) problem was first formally introduced by Blanchard et al.~\cite{blanchard2011generalizing} as a machine learning problem, while the term \emph{domain generalization} was later coined by Muandet et al.~\cite{muandet2013domain}. Unlike other related learning problems such as domain adaptation or transfer learning, DG considers the scenarios where target data is \emph{inaccessible} during model learning. In~\cite{blanchard2011generalizing}, the motivation behind DG originates from a medical application called automatic gating of flow cytometry data. The objective is to design algorithms to automate the process of classifying cells in patients' blood samples based on different properties, e.g., to distinguish between lymphocytes and non-lymphocytes. Such a technology is crucial in facilitating the diagnosis of the health of patients since manual gating is extremely time-consuming and requires domain-specific expertise. However, due to distribution shift between different patients' data, a classifier learned using data from historic patients does not generalize to new patients, and meanwhile, collecting new data for model fine-tuning is impractical, thus motivating research on the DG problem.

In computer vision, a seminal work done by Torralba and Efros~\cite{torralba2011unbiased} raised attention on the cross-domain generalization issue. They performed a thorough investigation into the cross-dataset generalization performance of object recognition models using six popular benchmark datasets. Their findings suggested that dataset biases, which are difficult to avoid, can lead to poor generalization performance. For example, as shown in~\cite{torralba2011unbiased}, a person classifier trained on Caltech101~\cite{fei2004learning} obtained a very low accuracy (11.8\%) on LabelMe~\cite{russell2008labelme}, though its same-dataset performance was near-perfect (99.6\%). Following~\cite{torralba2011unbiased}, Khosla et al.~\cite{khosla2012undoing} targeted the cross-dataset generalization problem in classification and detection tasks, and proposed to learn domain-specific bias vectors and domain-agnostic weight vectors based on support vector machine (SVM) classifiers.

\subsection{Problem Definition}\label{sec:bg;subsec:problem_def}

We first introduce some notations that will be used throughout this survey. Let $\mathcal{X}$ be the input (feature) space and $\mathcal{Y}$ the target (label) space, a \emph{domain} is defined as a joint distribution $P_{XY}$ on $\mathcal{X} \times \mathcal{Y}$.\footnote{We use $P_{XY}$ and $P(X,Y)$ interchangeably.} For a specific domain $P_{XY}$, we refer to $P_X$ as the marginal distribution on $X$, $P_{Y|X}$ the posterior distribution of $Y$ given $X$, and $P_{X|Y}$ the class-conditional distribution of $X$ given $Y$.

In the context of DG, we have access to $K$ similar but distinct source domains $\mathcal{S} = \{ S_k = \{ (x^{(k)}, y^{(k)}) \} \}_{k=1}^K$, each associated with a joint distribution $P_{XY}^{(k)}$. Note that $P_{XY}^{(k)} \neq P_{XY}^{(k')}$ with $k \neq k'$ and $k, k' \in \{1,...,K\}$. The goal of DG is to learn a predictive model $f: \mathcal{X} \to \mathcal{Y}$ using only source domain data such that the prediction error on an unseen target domain $\mathcal{T} = \{ x^{\mathcal{T}} \}$ is minimized. The corresponding joint distribution of the target domain $\mathcal{T}$ is denoted by $P_{XY}^{\mathcal{T}}$. Also, $P_{XY}^{\mathcal{T}} \neq P_{XY}^{(k)}$, $\forall k \in \{1,...,K\}$.

\keypoint{Multi-Source DG}
DG has typically been studied under two different settings, namely \emph{multi-source DG} and \emph{single-source DG}. The majority of research has been dedicated to the multi-source setting, which assumes multiple distinct but relevant domains are available (i.e., $K > 1$). As stated in~\cite{blanchard2011generalizing}, the original motivation for studying DG is to leverage multi-source data to learn representations that are invariant to different marginal distributions. This makes sense because without having access to the target data, it is challenging for a source-learned model to generalize well. As such, using multiple domains allows a model to discover stable patterns across source domains, which generalize better to unseen domains.

\keypoint{Single-Source DG}
In contrast, the single-source setting assumes training data is homogeneous, i.e., they are sampled from a single domain ($K = 1$). This problem is closely related to the topic of OOD robustness~\cite{hendrycks2019benchmarking,hendrycks2020augmix,tang2021selfnorm}, which investigates model robustness under image corruptions. Essentially, single-source DG methods do not require domain labels for learning and thus they are applicable to multi-source scenarios as well. In fact, most existing methods able to solve single-source DG do not distinguish themselves as a single- or a multi-source approach, but rather a more generic solution to OOD generalization, with experiments covering both single- and multi-source datasets~\cite{cvpr19jigen,bucci2020self,wang2019learning,huang2020self}.

\begin{table*}[ht!]
    \centering
    \tabstyle{5pt}
    \caption{Commonly used domain generalization datasets (categorized mainly based on applications).}
    \label{tab:datasets}
    \resizebox{\textwidth}{!}{
    \begin{tabular}{l r c l}
    \toprule
    & \textbf{\# samples} & \textbf{\# domains} & \textbf{Characterization of domain shift} \\
    \midrule
    \textbf{Handwritten digit recognition} & \\
    \quad- Rotated MNIST~\cite{ghifary2015domain} & 70,000 & 6 & Rotations (0, 15, 30, 45, 60 \& 75) \\
    \quad- Digits-DG~\cite{zhou2020learning} & 24,000 & 4 & MNIST~\cite{lecun1998mnist}, MNIST-M~\cite{ganin2015unsupervised}, SVHN~\cite{netzer2011svhn}, SYN~\cite{ganin2015unsupervised} \\
    \midrule
    \textbf{Object recognition} & \\
    \quad- VLCS~\cite{fang2013unbiased} & 10,729 & 4 & Caltech101~\cite{fei2004learning}, LabelMe~\cite{russell2008labelme}, PASCAL~\cite{everingham2010pascal}, SUN09~\cite{xiao2010sun} \\
    \quad- Office-31~\cite{saenko2010adapting} & 4,652 & 3 & Amazon, webcam, dslr \\
    \quad- OfficeHome~\cite{office_home} & 15,588 & 4 & Art, clipart, product, real \\
    \quad- PACS~\cite{li2017deeper} & 9,991 & 4 & Photo, art, cartoon, sketch \\
    \quad- DomainNet~\cite{iccv19domainnet} & 586,575 & 6 & Clipart, infograph, painting, quickdraw, real, sketch \\
    \quad- miniDomainNet~\cite{zhou2020domain} & 140,006 & 4 & Clipart, painting, real, sketch \\
    \quad- ImageNet-Sketch~\cite{wang2019learning} & 50,000 & 2 & Real vs sketch images \\
    \quad- VisDA-17~\cite{peng2017visda} & 280,157 & 2 & Synthetic vs real images \\
    \quad- CIFAR-10-C~\cite{hendrycks2019benchmarking} & 60,000 & - & Artificial corruptions \\
    \quad- CIFAR-100-C~\cite{hendrycks2019benchmarking} & 60,000 & - & Artificial corruptions \\
    \quad- ImageNet-C~\cite{hendrycks2019benchmarking} & $\approx$1.3M & - & Artificial corruptions \\
    \quad- ImageNet-R~\cite{hendrycks2021many} & 30k & - & Image style changes \\
    \quad- ImageNet-A~\cite{hendrycks2021natural} & 7,500 & - & Naturally adversarial examples \\
    \quad- TerraInc~\cite{beery2018recognition} & 24,788 & 4 & Geographical locations \\
    \quad- NICO++~\cite{zhang2022nico++} & 232.4k & 10 & Contexts (e.g., grass, water, winter, indoor, outdoor) \\
    \quad- Visual Decathlon~\cite{rebuffi2017learning} & 1,659,142 & 10 & Data sources (10 datasets) \\ 
    \midrule
    \textbf{Action recognition} & \\
    \quad- IXMAS~\cite{weinland2006free} & 1,650 & 5 & 5 camera views, 10 subjects (see~\cite{li2018mmdaae}) \\
    \quad- UCF-HMDB~\cite{soomro2012ucf101,kuehne2011hmdb} & 3,809 & 2 & Data sources (2 datasets) (see~\cite{yao2019adversarial}) \\
    \midrule
    \textbf{Semantic segmentation} & \\
    \quad- SYNTHIA~\cite{ros2016synthia} & 2,700 & 15 & 4 locations, 5 weather conditions (see~\cite{volpi2018generalizing}) \\
    \quad- GTA5-Cityscapes~\cite{richter2016playing,cordts2016cityscapes} & 29,966 & 2 & Synthetic vs real images \\
    \midrule
    \textbf{Person re-identification} & \\
    \quad- Market-Duke~\cite{zheng2015scalable,ristani2016performance} & 69,079 & 2 & Camera views, cities, streets, etc. \\
    \midrule
    \textbf{Face recognition} & \\
    \quad- Face~\cite{shi2020towards} & $>$5M & 9 & Data sources (9 datasets) \\
    \midrule
    \textbf{Face anti-spoofing} & \\
    \quad- COMI~\cite{zhang2012face,boulkenafet2017oulu,wen2015face,chingovska2012effectiveness} & $\approx$8,500 & 4 & Data sources (4 datasets) \\
    \midrule
    \textbf{Speech recognition} & \\
    \quad- Google Speech Command~\cite{sainath2015convolutional} & 65k & 1,888 & Speakers \\
    \midrule
    \textbf{Sentiment classification} & \\
    \quad- Amazon Reviews~\cite{blitzer2006domain} & $>$340k  & 4 & Books, DVD, electronics, kitchen appliances \\
    \midrule
    \textbf{WILDS}~\cite{wilds2020} (only show 3 out of 10 datasets here) & \\
    \quad- Camelyon17-WILDS~\cite{bandi2018detection} & 455,954 & 5 & Hospitals \\
    \quad- FMoW-WILDS~\cite{christie2018functional} & 523,846 & 80 & Time, geographical regions \\
    \quad- iWildCam-WILDS~\cite{beery2020iwildcam} & 203,029 & 323 & Camera traps \\
    \midrule
    \textbf{Medical imaging} & \\
    \quad- Multi-site Prostate MRI Segmentation~\cite{liu2020shape} & 116 & 6 & Clinical centers \\
    \quad-  Chest X-rays~\cite{mahajan2021domain} & - & 3 & Data sources (NIH~\cite{wang2017chestx}, ChexPert~\cite{irvin2019chexpert}, RSNA) \\
    \midrule
    \textbf{Reinforcement learning} & \\
    \quad- Coinrun~\cite{cobbe2019quantifying} & - & - & Scenes, difficulty levels \\
    \quad- OpenAI Procgen Benchmark~\cite{cobbe2020leveraging} & - & - & States, scenes, rewards, difficulty levels \\
    \bottomrule
    \end{tabular}
    }
\end{table*}

\subsection{Datasets and Applications}\label{sec:bg;subsec:data_app}
DG has been studied across many application areas including computer vision, speech recognition, medical imaging, and so on. Table~\ref{tab:datasets} summarizes the commonly used datasets based on different applications. Below we briefly discuss their basics.

\begin{figure*}[t]
    \centering
    \includegraphics[width=.95\textwidth]{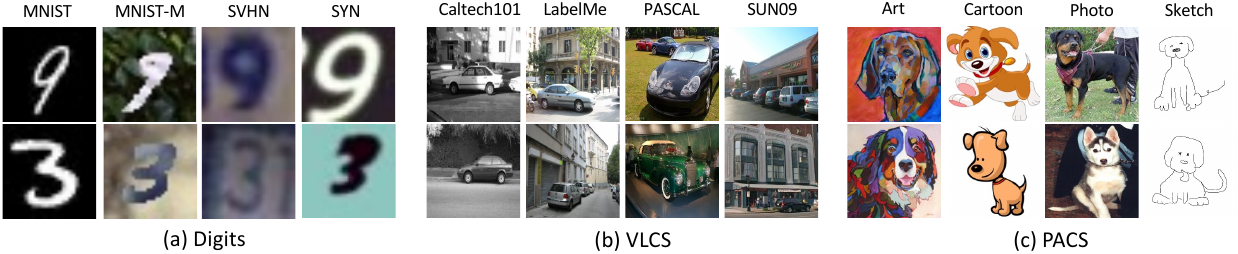}
    \caption{Example images from three domain generalization benchmarks manifesting different types of domain shift. In (a), the domain shift mainly corresponds to changes in font style, color and background. In (b), dataset-specific biases are clear, which are caused by changes in environment/scene and viewpoint. In (c), image style changes are the main reason for domain shift.
    }
    \label{fig:datasets}
\end{figure*}

\keypoint{Handwritten Digit Recognition}
The commonly used digit datasets include MNIST~\cite{lecun1998mnist}, MNIST-M~\cite{ganin2015unsupervised}, SVHN~\cite{netzer2011svhn}, and SYN~\cite{ganin2015unsupervised}. In general, these datasets differ in font style, stroke color, and background. MNIST contains images of handwritten digits. MNIST-M mixes MNIST's images with random color patches. SVHN comprises images of street view house numbers while SYN is a synthetic dataset. See Fig.~\ref{fig:datasets}(a) for some example images. Rotation has also been exploited to synthesize domain shift~~\cite{ghifary2015domain}.

\keypoint{Object Recognition}
has been the most common task in DG where the domain shift varies substantially across different datasets. In VLCS~\cite{fang2013unbiased} and Office-31~\cite{saenko2010adapting}, the domain shift is mainly caused by changes in environments or viewpoints. As exemplified in Fig.~\ref{fig:datasets}(b), the scenes in VLCS vary from urban to rural areas and the viewpoints are often biased toward either a side-view or a non-canonical view. Image style changes have also been commonly studied, such as PACS~\cite{li2017deeper} (see Fig.~\ref{fig:datasets}(c)), OfficeHome~\cite{office_home}, DomainNet~\cite{iccv19domainnet}, and ImageNet-Sketch~\cite{wang2019learning}. Other types of domain shift include synthetic-vs-real~\cite{fang2013unbiased}, artificial corruptions~\cite{hendrycks2019benchmarking}, and data sources~\cite{rebuffi2017learning}.

\keypoint{Action Recognition}
Learning generalizable representations is also crucial for video understanding like action recognition. IXMAS~\cite{weinland2006free} has been widely used as a cross-view action recognition benchmark~\cite{li2019episodic,li2018mmdaae}, which contains action videos collected from five different views. In addition to view changes, different subjects or environments (like indoor vs outdoor) can also create domain shift and lead to model failures.

\keypoint{Semantic Segmentation}
is critical to autonomous driving. Though this task has been greatly advanced by deep neural networks, the performance is still far from being satisfactory when deploying trained deep models in novel scenarios, such as new cities or unseen weather conditions~\cite{hoffman2018cycada}. Since it is impractical to collect data covering all possible scenarios, DG is pivotal in facilitating large-scale deployment of semantic segmentation systems. The SYNTHIA dataset~\cite{ros2016synthia} contains synthetic images of different locations under different weather conditions. Generalization from GTA5~\cite{richter2016playing} to real image datasets like Cityscapes~\cite{cordts2016cityscapes} has also been extensively studied~\cite{gong2018dlow}.

\keypoint{Person Re-Identification (Re-ID)}
plays a key role in security and surveillance applications. Person re-ID is essentially an instance retrieval task, aiming to match people across disjoint camera views (each seen as a distinct domain). Most existing methods in re-ID~\cite{sun2019learning,li2019scalable,chen2019abd,zhou2019osnet,chang2018multi} have been focused on the same-dataset setting, i.e., training and test are done on the same set of camera views, with performance almost reaching saturation. Recently, cross-dataset re-ID~\cite{zhou2021osnet,zhao2021learning,choi2020meta} has gained much attention: the objective is to generalize a model from source camera views to unseen target camera views, a more challenging but realistic setting. The domain shift often occurs in image resolution, viewpoint, lighting condition, background, etc.

\keypoint{Face Recognition}
has witnessed significant advances driven by deep learning in recent years~\cite{taigman2014deepface,sun2014deep,wen2016discriminative}. However, several studies~\cite{shi2020towards} have suggested that deep models trained even on large-scale datasets like MS-Celeb-1M~\cite{guo2016ms} suffer substantial performance drops when deployed in new datasets with previously unseen domains, such as low resolution~\cite{wolf2011face,kalka2018ijb,cheng2018low}, large variations in illumination/occlusion/head pose~\cite{klare2015pushing,maze2018iarpa,sengupta2016frontal}, or drastically different viewpoints~\cite{kemelmacher2016megaface}.

\keypoint{Face Anti-Spoofing}
aims to prevent face recognition systems from being attacked using fake faces~\cite{yang2014learn}, such as printed photos, videos or 3D masks. Conventional face anti-spoofing methods do not take into account distribution shift, making them vulnerable to unseen attack types~\cite{shao2019multi}. There is no specifically designed DG dataset for this task. A common practice is to combine several face anti-spoofing datasets for model training and do evaluation on an unseen dataset, e.g., using CASIA-MFSD~\cite{zhang2012face}, Oulu-NPU~\cite{boulkenafet2017oulu} and MSU-MFSD~\cite{wen2015face} as the sources and Idiap Replay-Attack~\cite{chingovska2012effectiveness} as the target.

\keypoint{Speech Recognition}
Since people speak differently (e.g., different tones or pitches) it is natural to regard each speaker as a domain~\cite{shankar2018generalizing}. The commonly used dataset is Google Speech Command~\cite{sainath2015convolutional}, which consists of 1,888 domains (speakers) and around 65,000 samples.

\keypoint{Sentiment Classification}
is a common task studied in natural language processing, which aims to classify opinions in texts as either positive or negative (hence a binary classification problem)~\cite{balaji2018metareg}. Amazon Reviews~\cite{blitzer2006domain} contains reviews for four categories (domains) of products: books, DVD, electronics and kitchen appliances.

\keypoint{The WILDS Benchmark}
has been recently introduced, with a goal to study distribution shift faced in the wild~\cite{wilds2020}. The benchmark contains a total of ten datasets, which cover a wide range of pattern recognition tasks, such as animal classification, cancer detection, molecule classification, and satellite imaging. Table~\ref{tab:datasets} shows three datasets from WILDS that have been commonly used by the DG community~\cite{shi2021gradient,robey2021model,yao2022improving}.

\keypoint{Medical Imaging}
DG is also critical to medical imaging where domain shift is often related to variations in clinical centers or patients~\cite{dou2019domain,liu2020shape}. Two commonly used medical imaging datasets are Multi-site Prostate MRI Segmentation~\cite{liu2020shape} and Chest X-rays~\cite{mahajan2021domain}, each containing data aggregated from multiple clinical centers with domain shift caused by, e.g., different scanners or acquisition protocols.

\keypoint{Reinforcement Learning (RL)}
has a dramatically different paradigm than supervised or unsupervised learning: RL aims to maximize rewards obtained through continuous interactions with an environment. Generalization in RL has been a critical issue where agents or policies learned in a training environment often suffer from overfitting and hence generalize poorly to unseen environments~\cite{mazoure2021improving,hansen2021generalization,deitke2020robothor,cobbe2019quantifying}. Domain shift in RL is mostly associated with environmental changes, such as different scenes, states, or even rewards. There is a large body of work focused on improving generalization in RL. We refer readers to \cite{kirk2021survey} for a more comprehensive survey in the topic of generalizable RL.

\begin{table*}[t]
\centering
\tabstyle{15pt}
\caption{Comparison between domain generalization and its related topics.}
\label{tab:compare_topics}
\resizebox{\textwidth}{!}{
    \begin{tabular}{l | x{.5cm} x{.5cm} | x{.5cm} x{.5cm} | x{.5cm} x{.5cm} | c}
    \toprule
     & \multicolumn{2}{c|}{$K$} & \multicolumn{2}{c|}{$P_{XY}^{\mathcal{S}}$ vs $P_{XY}^{\mathcal{T}}$} & \multicolumn{2}{c|}{$\mathcal{Y}_S$ vs $\mathcal{Y}_T$} & \multirow{2}{*}{Access to $P_{X}^{\mathcal{T}}$?} \\
     & $=1$ & $>1$ & $=$ & $\neq$  & $=$ & $\neq$ & \\
    \midrule
    Supervised Learning & \cmark &  & \cmark &  & \cmark &  &  \\
    Multi-Task Learning &  & \cmark & \cmark &  & \cmark &  &  \\
    Transfer Learning & \cmark & \cmark &  & \cmark &  & \cmark & \cmark \\
    Zero-Shot Learning & \cmark & & & \cmark & & \cmark & \\
    Domain Adaptation & \cmark & \cmark &  & \cmark & \cmark & \cmark & \cmark \\
    Test-Time Training & \cmark & \cmark & & \cmark & \cmark & & \cmark$^\dagger$ \\
    \textbf{Domain Generalization} & \cmark & \cmark &  & \cmark & \cmark & \cmark &  \\
    \bottomrule
    \end{tabular}
}
\begin{flushleft}
\footnotesize
\itshape
$K$: number of source domains/tasks. $P_{XY}^{S/T}$: source/target joint distribution. $\mathcal{Y}_{S/T}$: source/target label space. $P_X^{\mathcal{T}}$: target marginal. $\dagger$: Limited in quantities, like a single example or mini-batch.
\end{flushleft}
\end{table*}

\subsection{Evaluation}\label{sec:bg;subsec:eval}
Evaluation of DG algorithms often follows the \emph{leave-one-domain-out} rule~\cite{li2017deeper}: given a dataset containing at least two distinct domains, one or multiple of them are used as source domain(s) for model training while the rest are treated as target domain(s); a model learned from the source domain(s) is directly tested in the target domain(s) without any form of adaptation. Two problem scenarios have been studied: single- vs multi-source DG. It is worth noting that some datasets contain label shift, meaning that the label space between source and target changes (termed heterogeneous DG~\cite{feature_critic}). For instance, in the problem of person re-ID, identities between training and test are different; in this case the source-learned representation is directly used for image matching.

\keypoint{Evaluation Metrics}
Two metrics have been commonly adopted, namely \emph{average} and \emph{worst-case} performance. The former concerns about the average performance in held-out domains, which is used in most domain shift scenarios. In contrast, the latter focuses on the worst performance among held-out domains, which is often used in the case of sub-population shift~\cite{sagawa2019distributionally} and has been widely adopted by the causal inference community~\cite{krueger2021out} as well as some datasets in the WILDS benchmark~\cite{wilds2020}.

\keypoint{Model Selection}
concerns about which model (checkpoint), architecture or hyper-parameters to choose for evaluation, which has recently been identified by \cite{gulrajani2020search} as a crucial step in the evaluation pipeline. As summarized in \cite{gulrajani2020search}, there are three model selection criteria: \romannum{1}) \emph{Training-domain validation}, which holds out a subset of training data for model selection; \romannum{2}) \emph{Leave-one-domain-out validation}, which keeps one source domain for model selection; \romannum{3}) \emph{Test-domain validation (oracle)}, which performs model selection using a random subset of test domain data. As suggested by \cite{gulrajani2020search}, the last criterion would lead to overly optimistic or pessimistic results and thus should be used with care. Another important lesson from \cite{gulrajani2020search} is that specially designed DG methods often perform similarly with the plain model (known as Empirical Risk Minimization) when using larger neural networks and an extensive search of hyper-parameters. Therefore, it is suggested that future evaluation should cover different neural network architectures and ensure comparison is made using the same model selection criterion.

\subsection{Related Topics}\label{sec:bg;subsec:related_topics}
In this section, we discuss the relations between DG and its related topics, and clarify their differences. See Table~\ref{tab:compare_topics} for an overview.

\keypoint{Supervised Learning}
generally aims to learn an input-output mapping by minimizing the following risk: $\mathbb{E}_{(x, y) \sim \hat{P}_{XY}} \ell( f(x), y )$, where $\hat{P}_{XY}$ denotes the empirical distribution rather than the real data distribution $P_{XY}$, which is inaccessible. The hope is that once the loss is minimized, the learned model can work well on data generated from $P_{XY}$, which heavily relies on the i.i.d.~assumption. The crucial difference between SL and DG is that in the latter training and test data is drawn from different distributions, thus violating the i.i.d.~assumption. DG is arguably a more practical setting in real-world applications~\cite{torralba2011unbiased}.

\keypoint{Multi-Task Learning (MTL)}
The goal of MTL is to simultaneously learn multiple related tasks ($K>1$) using a single model~\cite{yang2017deep,mallya2018piggyback,liu2019end,guo2020learning,sun2020adashare}. As shown in Table~\ref{tab:compare_topics}, MTL aims to make a model perform well on the same set of tasks that the model was trained on ($\mathcal{Y}_S = \mathcal{Y}_T$), whereas DG aims to generalize a model to unseen data distributions ($P_{XY}^{\mathcal{S}} \neq P_{XY}^{\mathcal{T}}$). Though being different in terms of the problem setup, the MTL paradigm has been exploited in some DG methods, notably for those based on self-supervised learning~\cite{cvpr19jigen,wang2020learning,albuquerque2020improving}. Intuitively, MTL benefits from the effect of regularization brought by parameter sharing~\cite{yang2017deep}, which may in part explain why the MTL paradigm works for DG.

\keypoint{Transfer Learning (TL)}
aims to transfer the knowledge learned from one (or multiple) problem/domain/task to a different but related one~\cite{pan2009survey}. A well-known TL example in contemporary deep learning is fine-tuning: first pre-train deep neural networks on large-scale datasets, such as ImageNet~\cite{deng2009imagenet} for vision models or BooksCorpus~\cite{zhu2015aligning} for language models; then fine-tune them on downstream tasks~\cite{girshick2014rich}. Given that pre-trained deep features are highly transferable, as shown in several studies~\cite{yosinski2014transferable,donahue2014decaf}, a couple of recent DG works~\cite{chen2020automated,chen2021contrastive} have researched how to preserve the transferable features learned via large-scale pre-training when learning new knowledge from source synthetic data for synthetic-to-real applications. As shown in Table~\ref{tab:compare_topics}, a key difference between TL and DG lies in whether the target data is used. In TL, the target data is required for model fine-tuning for new downstream tasks, whereas in DG we assume to have no access to the target data, thus focusing more on model generalization. Nonetheless, TL and DG share some similarities: the target distribution in both TL and DG is different from the source distribution; in terms of label space, TL mainly concerns disjoint label space, whereas DG considers both cases, i.e., same label space for homogeneous DG and disjoint label space for heterogeneous DG.

\keypoint{Zero-Shot Learning (ZSL)}
is related to DG in the sense that the goal in both problems is to deal with unseen distributions. Differently, distribution shift in ZSL is mainly caused by label space changes~\cite{lampert2014attribute,zhou2021coop,zhou2022cocoop}, i.e., $P_Y^{\mathcal{T}} \neq P_Y^{\mathcal{S}}$, since the task is to recognize new classes, except for generalized ZSL~\cite{chao2016empirical} which considers both new and old classes at test time; while in DG, domain shift mostly results from covariate shift~\cite{muandet2013domain}, i.e., only the marginal distribution changes ($P_X^{\mathcal{T}} \neq P_X^{\mathcal{S}}$).\footnote{It is worth mentioning that a recent ZSL work~\cite{mancini2020towards} has studied ZSL+DG, i.e., distribution shift occurs in both $P_Y$ and $P_X$, which is analogous to heterogeneous DG.} To recognize unseen classes in ZSL, a common practice is to learn a mapping between the input image space and the attribute space~\cite{xian2018zero} since the label space is disjoint between training and test data. Interestingly, attributes have also been exploited in DG for learning domain-generalizable representations~\cite{gan2016learning}.

\keypoint{Domain Adaptation (DA)}
is the closest topic to DG and has been extensively studied in the literature~\cite{gong2012geodesic,long2015learning,long2016unsupervised,balaji2019normalized,saito2018maximum,ganin2015unsupervised,kang2019contrastive,hoffman2018cycada,lu2020stochastic,liu2020open}. Both DA and DG aim to tackle the domain shift problem (i.e., $P_{XY}^{\mathcal{S}} \neq P_{XY}^{\mathcal{T}}$) encountered in new test environments. Differently, DA assumes the availability of sparsely labeled~\cite{kulis2011you} or unlabeled~\cite{gong2012geodesic} target data for model adaptation, hence having access to the marginal $P_X^{\mathcal{T}}$. Though there exist different variants of DA where some methods do not explicitly use target data during training, such as zero-shot DA~\cite{peng2018zero} that exploits task-irrelevant but target domain-relevant data (equivalent to accessing the marginal), their main idea remains unchanged, i.e., to leverage additional data that expose information related to the target domain. As shown in Table~\ref{tab:compare_topics}, research in DA shares some commonalities with DG, such as single-~\cite{gong2012geodesic} and multi-source~\cite{iccv19domainnet} DA, and heterogeneous DA~\cite{panareda2017open,liu2020open,cao2018partial,you2019universal}.

\keypoint{Test-Time Training (TTT)}
also called test-time adaptation~\cite{wang2020tent}, blurs the boundary between DA and DG. As shown in Table~\ref{tab:compare_topics}, TTT is related to both DA and DG because TTT also deals with the domain shift problem. One one hand, TTT (mostly) bears a resemblance with source-free DA~\cite{kundu2020universal}---both assume source data is inaccessible after model training. On the other hand, TTT differs from DA in that only a single~\cite{zhang2021memo} or mini-batch~\cite{iwasawa2021test} test data is used for model \emph{tuning}, which is often done in an online manner~\cite{sun2020test} and of course, without human supervision. It is also worth mentioning that datasets used in the TTT community largely overlap with those in DG, such as CIFAR-C~\cite{hendrycks2019benchmarking} and ImageNet-C~\cite{hendrycks2019benchmarking}. In terms of performance, TTT likely outperforms DG~\cite{iwasawa2021test} due to the use of test data for parameter update, but is limited to scenarios where deployment devices have sufficient compute power. Moreover, TTT might not fit realtime applications if the tuning time is too ``long.''

\begin{table*}
    \centering
    \tabstyle{5pt}
    \caption{Categorization of domain generalization (DG) methods.}
    \label{tab:categorization_dg_methods}
    \resizebox{\textwidth}{!}{
        \begin{tabular}{l c l}
        \toprule
        & \textbf{Domain labels} & \textbf{References} \\
        \midrule
        \multicolumn{3}{l}{\textbf{Domain Alignment} (\S~\ref{sec:methods;subsec:alignment})} \\
        \quad- Minimizing Moments & \cmark & \cite{muandet2013domain,erfani2016robust,ghifary2017scatter,li2018domain,jin2020feature,hu2020domain} \\
        \quad- Minimizing Contrastive Loss & \cmark & \cite{motiian2017unified,yoon2019generalizable,mahajan2020domain} \\
        \quad- Minimizing the KL Divergence & \cmark & \cite{wang2020respecting,li2020domain} \\
        \quad- Minimizing Maximum Mean Discrepancy & \cmark & \cite{li2018mmdaae} \\
        \quad- Domain-Adversarial Learning & \cmark & \cite{li2018ciddg,shao2019multi,rahman2020correlation,albuquerque2019generalizing,deng2020representation,matsuura2020domain,jia2020single,akuzawa2019adversarial,aslani2020scanner,zhao2020domain} \\
        \midrule
        \textbf{Meta-Learning} (\S~\ref{sec:methods;subsec:meta_learning}) & \cmark & \cite{li2018learning,li2020sequential,balaji2018metareg,li2019episodic,du2020learning,liu2020shape,feature_critic,dou2019domain,zhao2021learning,choi2020meta,du2021metanorm,wang2020meta} \\
        \midrule
        \multicolumn{3}{l}{\textbf{Data Augmentation} (\S~\ref{sec:methods;subsec:data_aug})} \\
        \quad- Image Transformations & \xmark & \cite{volpi2019addressing,otalora2019staining,chen2020improving,zhang2020generalizing,shi2020towards} \\
        \quad- Task-Adversarial Gradients & \xmark & \cite{volpi2018generalizing,qiao2020learning,sinha2017certifying} \\
        \quad- Domain-Adversarial Gradients & \cmark & \cite{shankar2018generalizing} \\
        \quad- Random Augmentation Networks & \xmark & \cite{xu2021robust} \\
        \quad- Off-the-Shelf Style Transfer Models & \cmark & \cite{somavarapu2020frustratingly,borlino2021rethinking,zhou2021stylematch,yue2019domain} \\
        \quad- Learnable Augmentation Networks & \cmark & \cite{zhou2020learning,zhou2020deep,carlucci2019hallucinating} \\
        \quad- Feature-Based Augmentation & \cmark & \cite{zhou2021mixstyle,zhou2021mixstylenn,mancini2020towards} \\
        \midrule
        \multicolumn{3}{l}{\textbf{Ensemble Learning} (\S~\ref{sec:methods;subsec:ensemble})} \\
        \quad- Exemplar-SVMs & \xmark & \cite{xu2014exploiting,niu2015multiview,niu2015visual} \\
        \quad- Domain-Specific Neural Networks & \cmark & \cite{ding2017deep,zhou2020domain,d2018domain,mancini2018best,wang2020dofe} \\
        \quad- Domain-Specific Batch Normalization & \cmark & \cite{seo2020learning,liu2020ms,segu2020batch,mancini2018robust} \\
        \quad- Weight Averaging & \xmark & \cite{cha2021domain} \\
        \midrule
        \multicolumn{3}{l}{\textbf{Self-Supervised Learning} (\S~\ref{sec:methods;subsec:self_supervised})} \\
        \quad- Single Pretext Task & \xmark & \cite{cvpr19jigen,wang2020learning,ghifary2015domain,maniyar2020zero} \\
        \quad- Multiple Pretext Tasks & \xmark & \cite{albuquerque2020improving,bucci2020self} \\
        \midrule
        \multicolumn{3}{l}{\textbf{Learning Disentangled Representations} (\S~\ref{sec:methods;subsec:disentangled})} \\
        \quad- Decomposition & \cmark & \cite{khosla2012undoing,li2017deeper,chattopadhyay2020learning,piratla2020efficient} \\
        \quad- Generative Modeling & \cmark & \cite{ilse2019diva,wang2020cross} \\
        \midrule
        \textbf{Regularization Strategies} (\S~\ref{sec:methods;subsec:reg}) & \xmark & \cite{wang2019learning,huang2020self} \\
        \midrule
        \multicolumn{3}{l}{\textbf{Reinforcement Learning} (\S~\ref{sec:methods;subsec:rl})} \\
        \quad- Data augmentation & \xmark & \cite{laskin2020reinforcement,kostrikov2020image,hansen2021generalization,tobin2017domain,lee2019network,zhou2021mixstyle} \\
        \quad- Self-Supervision & \xmark & \cite{yarats2021improving,laskin2020curl} \\
        \bottomrule
        \end{tabular}
    }
    \begin{flushleft}
        \footnotesize
        \itshape
        Note that methods requiring domain labels can only be applied to multi-source DG while those not requiring domain labels are applicable to both multi- and single-source DG.
    \end{flushleft}
\end{table*}

\section{Methodologies: A Survey}\label{sec:methods}
Numerous domain generalization (DG) methods have been proposed in the past ten years, and the majority of them are designed for multi-source DG, despite some methods not explicitly requiring domain labels for learning and thus suitable for single-source DG as well. In this section, we categorize existing DG methods into several groups based on their methodologies and motivations behind their design. Within each group, we further discuss different variants and indicate whether domain labels are required for learning to differentiate their uses---those requiring domain labels can only be applied to multi-source DG while those not requiring domain labels are applicable to both single- and multi-source DG. See Table~\ref{tab:categorization_dg_methods} for an overview.

\begin{figure}[t]
	\centering
	\includegraphics[width=.8\columnwidth]{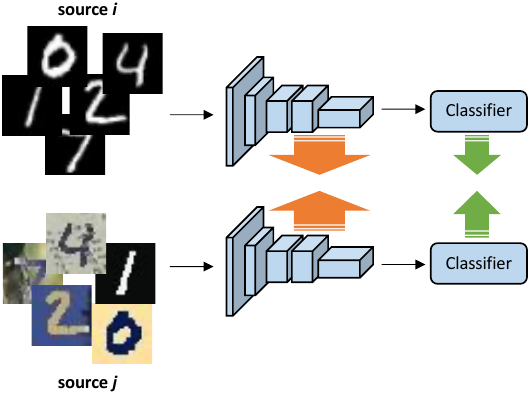}
	\caption{Domain alignment is commonly applied to a pair of source domains, either in the feature space (orange arrows) or the classifier's output (green arrows), or both.}
	\label{fig:domain_alignment}
\end{figure}

\subsection{Domain Alignment}\label{sec:methods;subsec:alignment}
Most existing DG approaches belong to the category of domain alignment~\cite{muandet2013domain,motiian2017unified,wang2020respecting,li2018mmdaae,li2018ciddg,shao2019multi}, where the central idea is to minimize the difference among source domains for learning domain-invariant representations (see Fig.~\ref{fig:domain_alignment}). The motivation is straightforward: features that are invariant to the source domain shift should also be robust to any unseen target domain shift. Domain alignment has been applied in many DG applications, e.g., object recognition~\cite{li2018ciddg,ghifary2015domain}, action recognition~\cite{li2018mmdaae}, face anti-spoofing~\cite{shao2019multi,jia2020single}, and medical imaging analysis~\cite{aslani2020scanner,li2020domain}. Domain labels are required for domain alignment methods.

To measure the distance between distributions and thereby achieve alignment, there are a wide variety of statistical distance metrics for us to borrow, such as the simple $\ell_2$ distance, $f$-divergences, or the more sophisticated Wasserstein distance~\cite{villani2008optimal}. However, designing an effective domain alignment method is a non-trivial task because one needs to consider \emph{what to align} and \emph{how to align}. In the following sections, we analyze the existing alignment-based DG methods from these two aspects.

\subsubsection{What to Align}
Recall that a domain is modeled by a joint distribution $P(X,Y)$ (see \S~\ref{sec:bg;subsec:problem_def} for the background), we can decompose it into
\begin{align}
P(X,Y) &= P(Y|X) P(X), \label{eq:p(y|x)p(x)} \\
       &= P(X|Y) P(Y). \label{eq:p(x|y)p(y)}
\end{align}

A common assumption in DG is that distribution shift only occurs in the marginal $P(X)$ while the posterior $P(Y|X)$ remains relatively stable~\cite{muandet2013domain} (see Eq.~\eqref{eq:p(y|x)p(x)}). Therefore, numerous domain alignment methods have been focused on aligning the marginal distributions of source domains~\cite{muandet2013domain,erfani2016robust,li2018mmdaae,ghifary2017scatter}.

From a causal learning perspective~\cite{scholkopf2012causal}, it is valid to align $P(X)$ only when $X$ is the cause of $Y$. In this case, $P(Y|X)$ is not coupled with $P(X)$ and thus remains stable when $P(X)$ varies. However, it is also possible that $Y$ is the cause of $X$, and as a result, shift in $P(X)$ will also affect $P(Y|X)$. Therefore, some domain alignment methods~\cite{li2018domain,hu2020domain,li2018ciddg} proposed to instead align the class-conditional $P(X|Y)$, assuming that $P(Y)$ does not change (see Eq.~\eqref{eq:p(x|y)p(y)}). For example, Li et al.~\cite{li2018domain} learned a feature transformation by minimizing for all classes the variance of class-conditional distributions across source domains. To allow $P(Y)$ to change along with $P(X|Y)$, i.e., heterogeneous DG, Hu et al.~\cite{hu2020domain} relaxed the assumption made in~\cite{li2018domain} by removing the minimization constraint on marginal distributions and proposed several discrepancy measures to learn generalizable features.

Since the posterior $P(Y|X)$ is what we need at test time, Wang et al.~\cite{wang2020respecting} introduced hypothesis invariant representations, which are obtained by directly aligning the posteriors within each class regardless of domains via the Kullback–Leibler (KL) divergence.

\subsubsection{How to Align}
Having discussed what to align in the previous section, here we turn to the exact techniques used in the DG literature for distribution alignment.

\keypoint{Minimizing Moments}
Moments are parameters used to measure a distribution, such as mean (1st-order moment) and variance (2nd-order moment) calculated over a population. Therefore, to achieve invariance between source domains, one can learn a mapping function (e.g., a simple projection matrix~\cite{ghifary2017scatter} or a complex non-linear function modeled by deep neural networks~\cite{jin2020feature}) with an objective of minimizing the moments of the transformed features between source domains, in terms of variance~\cite{muandet2013domain,erfani2016robust} or both mean and variance~\cite{ghifary2017scatter,li2018domain,jin2020feature,hu2020domain}.

\keypoint{Minimizing Contrastive Loss}
is another option for reducing distribution mismatch~\cite{motiian2017unified,yoon2019generalizable,mahajan2020domain}, which takes into account the semantic labels. There are two key design principles. The first is about how to construct the anchor group, the positive group (same class as the anchor but from different domains) and the negative group (different class than the anchor). The second is about the formulation of the distance function (e.g., using $\ell_2$~\cite{motiian2017unified} or softmax~\cite{mahajan2020domain}). The objective is to pull together the anchor and the positive groups, while push away the anchor and the negative groups.

\keypoint{Minimizing the KL Divergence}
As a commonly used distribution divergence measure, the KL divergence has also been employed for domain alignment~\cite{wang2020respecting,li2020domain}. In~\cite{wang2020respecting}, domain-agnostic posteriors within each class are aligned via the KL divergence. In~\cite{li2020domain}, the KL divergence is used to force all source domain features to be aligned with a Gaussian distribution.

\keypoint{Minimizing Maximum Mean Discrepancy (MMD)}
The MMD distance~\cite{gretton2012kernel} measures the divergence between two probability distributions by first mapping instances to a reproducing kernel Hilbert space (RKHS) and then computing the distance based on their mean. Using the autoencoder architecture, Li et al.~\cite{li2018mmdaae} minimized the MMD distance between source domain distributions on the hidden-layer features, and meanwhile, forced the feature distributions to be similar to a prior distribution via adversarial learning~\cite{goodfellow2014generative}.

\keypoint{Domain-Adversarial Learning}
Different from explicit distance measures like the MMD, adversarial learning~\cite{goodfellow2014generative} formulates the distribution minimization problem through a minimax two-player game. Initially proposed by Goodfellow et al.~\cite{goodfellow2014generative}, adversarial learning was used to train a generative model, which takes as input random noises and generates photorealistic images. This is achieved by learning a discriminator to distinguish between real and the generated fake images (i.e., minimizing the binary classification loss), while encouraging the generator to fool the discriminator (i.e., maximizing the binary classification loss). In particular, the authors in~\cite{goodfellow2014generative} theoretically justified that generative adversarial learning is equivalent to minimizing the Jensen-Shannon divergence between the real distribution and the generated distribution. Therefore, it is natural to use adversarial learning for distribution alignment, which has already been extensively studied in the domain adaptation area for aligning the source-target distributions~\cite{ganin2015unsupervised,tzeng2017adversarial,zhang2018collaborative,long2018conditional}.

In DG, adversarial learning is performed between source domains to learn source domain-agnostic features that are expected to work in novel domains~\cite{li2018ciddg,shao2019multi,rahman2020correlation,albuquerque2019generalizing,deng2020representation}. Simply speaking, the learning objective is to make features confuse a domain discriminator, which can be implemented as a multi-class domain discriminator~\cite{matsuura2020domain,akuzawa2019adversarial,aslani2020scanner}, or a binary domain discriminator in a per-domain basis~\cite{li2018ciddg,shao2019multi}. Typically, the learning steps alternate between the feature generator and the domain discriminator(s)~\cite{li2018ciddg}. However, one can simplify the process to achieve single-step update by using the gradient-reversal layer~\cite{ganin2015unsupervised} to flip the sign of the gradients back-propagated from the domain discriminator(s)~\cite{jia2020single}.

To enhance domain alignment, researchers have also combined domain-adversarial learning with explicit distance measures like moments minimization~\cite{rahman2020correlation}, or with some regularization constraints such as entropy~\cite{zhao2020domain}.

\keypoint{Multi-Task Learning}
has also been explored for domain alignment~\cite{ghifary2015domain,maniyar2020zero}. Different from directly minimizing the distribution divergence, MTL facilitates the learning of generic features by parameter sharing~\cite{yang2017deep}. This is easy to understand: in order to simultaneously deal with different tasks the features have to be generic enough. In~\cite{ghifary2015domain}, the authors proposed a denoising autoencoder architecture (later employed in~\cite{maniyar2020zero}) where the encoder is shared but the decoder is split into domain-specific branches, each connected to a reconstruction task. The model was trained with two objectives, one being self-domain reconstruction while the other being cross-domain reconstruction, which aim to force the hidden representations to be as generic as possible.

Domain alignment is still a popular research direction in DG. This idea has also been extensively studied in the domain adaptation (DA) literature~\cite{ganin2015unsupervised,long2016unsupervised,long2015learning,lee2019sliced,ganin2016domain}, but with a rigorous theoretical support~\cite{ben2010theory}. In particular, the DA theory introduced in~\cite{ben2010theory} suggested that minimizing the distribution divergence between source and target has a huge impact on lowering the upper-bound of the target error. However, in DG we cannot access the target data and therefore, the alignment is performed only among source domains. This inevitably raises a question of whether a representation learned to be invariant to source domain shift is guaranteed to generalize to an unseen domain shift in the target data. To solve this concern, one can focus on developing novel theories to explain how alignment in source domains improves generalization in unseen domains.

\begin{figure}[t]
    \centering
    \includegraphics[width=\columnwidth]{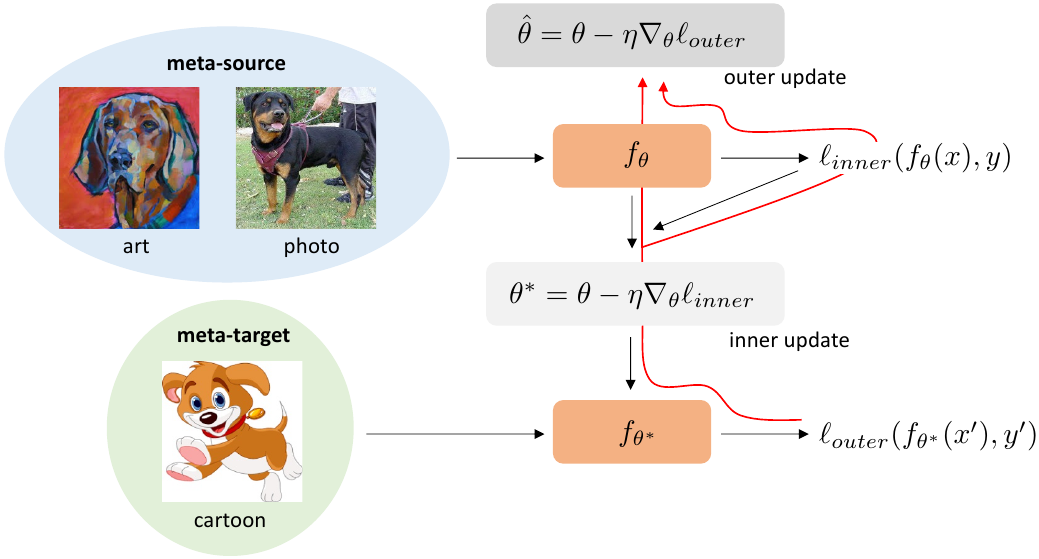}
    \caption{A commonly used meta-learning paradigm~\cite{li2018learning} in domain generalization. The source domains (i.e., art, photo and cartoon from PACS~\cite{li2017deeper}) are divided into disjoint meta-source and meta-target domains. The outer learning, which simulates domain shift using the meta-target data, back-propagates the gradients all the way back to the base parameters such that the model learned by the inner algorithm with the meta-source data improves the outer objective. The red arrows in this figure denote the gradient flow through the second-order differentiation.}
    \label{fig:meta_learning_paradigm}
\end{figure}

\subsection{Meta-Learning}\label{sec:methods;subsec:meta_learning}
Meta-learning has been a fast growing area with applications to many machine learning and computer vision problems~\cite{finn2017model,li2018learning,liu2020shape,zhao2021learning,li2019episodic}. Also known as learning-to-learning, meta-learning aims to learn from episodes sampled from related tasks to benefit future learning (see~\cite{hospedales2020meta} for a comprehensive survey on meta-learning). The meta-learning paper most related to DG is MAML~\cite{finn2017model}, which divides training data into meta-train and meta-test sets, and trains a model using the meta-train set in such a way to improve the performance on the meta-test set. The MAML-style training usually involves a second-order differentiation through the update of the base model, thus posing issues on efficiency and memory consumption for large neural network models~\cite{hospedales2020meta}. In~\cite{finn2017model}, MAML was used for parameter initialization, i.e., to learn an initialization state that is only a few gradient steps away from the solution to the target task.

The motivation behind applying meta-learning to DG is to expose a model to domain shift during training with a hope that the model can better deal with domain shift in unseen domains. Existing meta-learning DG methods can only be applied to multi-source DG where domain labels are provided.

There are two components that need to be carefully designed, namely \emph{episodes} and \emph{meta-representation}. Specifically, episodes construction concerns how each episode should be constructed using available samples, while meta-representation answers the question of what to meta-learn.

\keypoint{Episodes Construction}
Most existing meta-learning-based DG methods~\cite{li2020sequential,balaji2018metareg,feature_critic,du2020learning,liu2020shape,dou2019domain,zhao2021learning,choi2020meta,du2021metanorm} followed the learning paradigm proposed in~\cite{li2018learning}---which is the first method applying meta-learning to DG. Specifically, source domains are divided into non-overlapping \emph{meta-source} and \emph{meta-target} domains to simulate domain shift. The learning objective is to update a model using the meta-source domain(s) in such a way that the test error on the meta-target domain can be reduced, which is often achieved by bi-level optimization. See Fig.~\ref{fig:meta_learning_paradigm} for a graphical representation.

\keypoint{Meta-Representation}
is a term defined in~\cite{hospedales2020meta} to represent the model parameters that are meta-learned. Most deep learning methods meta-learned the entire neural network models~\cite{li2018learning,li2020sequential,dou2019domain}. Balaji et al.~\cite{balaji2018metareg} instead proposed to meta-learn the regularization parameters. In~\cite{du2020learning}, a stochastic neural network is meta-learned to handle uncertainty. In~\cite{liu2020shape}, an MRI segmentation model is meta-learned, along with two shape-aware losses to ensure compactness and smoothness in the segmentation results. Batch normalization layers are meta-learned in~\cite{zhao2021learning,choi2020meta,du2021metanorm} to cope with the training-test discrepancy in CNN feature statistics.

Overall, meta-learning is a promising direction to work on given its effectiveness in not only DG but also a wide range of applications like few-shot classification~\cite{finn2017model}, object detection~\cite{perez2020incremental} and image generation~\cite{gordon2019meta}. However, meta-learning in DG still suffers the same issue with that in domain alignment---a representation is only learned to be robust under source domain shift (simulated by meta-source and meta-target domains). Such an issue could be aggravated if the source domains are limited in terms of diversity. As observed from recent work~\cite{zhou2020learning,cha2021domain}, both meta-learning and domain alignment methods are underperformed by methods based on directly augmenting the source training data---a topic that will be visited later. One might alleviate the generalization issue in meta-learning, as well as in domain alignment, by combining them with data augmentation. Moreover, advances may also be achieved by designing novel meta-learning algorithms in terms of meta-representation, meta-optimizer, and/or meta-objective.\footnote{These terms are defined in~\cite{hospedales2020meta}.}

\begin{figure*}[t]
    \centering
    \includegraphics[width=.9\textwidth]{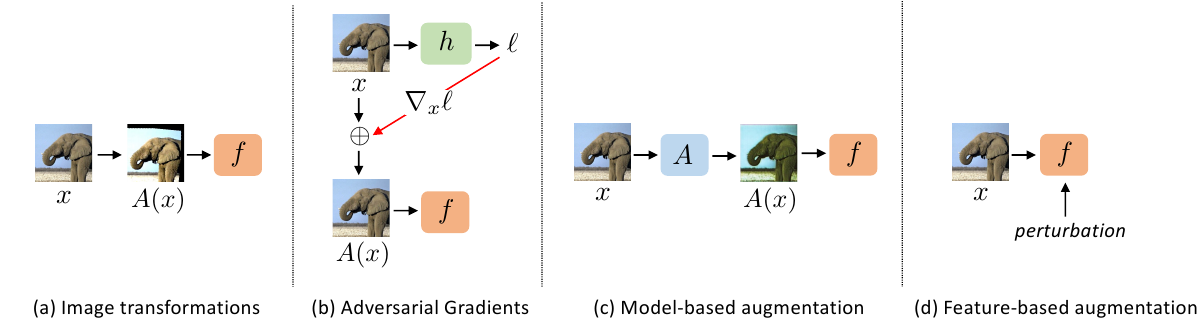}
    \caption{
    Based on the formulation of the transformation $A(\cdot)$, existing data augmentation methods can be categorized into four groups. \textbf{a)} The first group enhances the generalization of the classifier $f$ by applying hand-engineered image transformations like random crop or color augmentation to simulating domain shift. \textbf{b)} The second group is based on adversarial gradients obtained from either a category classifier ($h = f$) or a domain classifier. \textbf{c)} The third group models $A(\cdot)$ using neural networks, such as random CNNs~\cite{xu2021robust}, an off-the-shelf style transfer model~\cite{zhou2021stylematch}, or a learnable image generator~\cite{zhou2020learning}. \textbf{d)} The final group injects perturbation into intermediate features in the task model.
    }
    \label{fig:data_aug_pipelines}
\end{figure*}

\begin{figure}[t]
    \centering
    \includegraphics[width=.75\columnwidth]{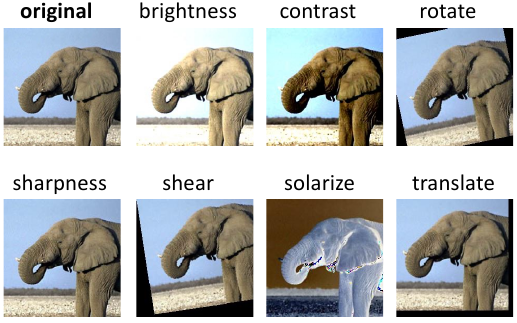}
    \caption{Common image transformations used as data augmentation in domain generalization~\cite{volpi2019addressing,otalora2019staining,chen2020improving,zhang2020generalizing}.}
    \label{fig:image_transformations}
\end{figure}

\subsection{Data Augmentation}\label{sec:methods;subsec:data_aug}
Data augmentation has been a common practice to regularize the training of machine learning models to avoid over-fitting and improve generalization~\cite{goodfellow2016deep}, which is particularly important for over-parameterized deep neural networks. The basic idea in data augmentation is to augment the original $(x, y)$ pairs with new $(A(x), y)$ pairs where $A(\cdot)$ denotes a transformation, which is typically label-preserving. Not surprisingly, given the advantages of data augmentation, it has been extensively studied in DG where $A(\cdot)$ is usually seen as a way of simulating domain shift and the design of $A(\cdot)$ is key to performance.

Based on how $A(\cdot)$ is formulated, data augmentation methods generally fall into four groups. See Fig.~\ref{fig:data_aug_pipelines} for an overview. Below we provide more detailed reviews, with a more fine-grained categorization where adversarial gradients are divided into task-adversarial gradients and domain-adversarial gradients; and model-based augmentation is further split into three sub-groups: random augmentation networks, off-the-shelf style transfer models, and learnable augmentation networks.

\keypoint{Image Transformations}
This type of approach exploits traditional image transformations, such as random flip, rotation and color augmentation. Fig.~\ref{fig:image_transformations} visualizes some effects of transformations. Though using image transformations does not require domain labels during learning, the selection of transformations is usually problem-specific. For example, for object recognition where image style changes are the main domain shift, one can choose transformations that are more related to color intensity changes, such as \texttt{brightness}, \texttt{contrast} and \texttt{solarize} in Fig.~\ref{fig:image_transformations}. To avoid manual picking, one can design a searching mechanism to search for the optimal set of transformations that best fit the target problem. An example is~\cite{volpi2019addressing} where the authors proposed an evolution-based searching algorithm and used a worst-case formulation to make the transformed images deviate as much as possible from the original image distribution. One can also select transformations according to the specific downstream task. For instance, \cite{shi2020towards} addressed the universal feature learning problem in face recognition by synthesizing meaningful variations such as lowering image resolution, adding occlusions and changing head poses.

Traditional image transformations have been shown very effective in dealing with domain shift in medical images~\cite{otalora2019staining,chen2020improving,zhang2020generalizing}. This makes sense because image transformations can well simulate changes in color and geometry caused by device-related domain shift, such as using different types of scanners in different medical centers. However, image transformations can be limited in some applications as they might cause label shift, such as digit recognition or optical character recognition where the horizontal/vertical flip operation is infeasible. Therefore, transformations should be carefully chosen to not conflict with the downstream task.

\keypoint{Task-Adversarial Gradients}
Inspired by adversarial attacks~\cite{goodfellow2015explaining,szegedy2014intriguing}, several data augmentation methods are based on using adversarial gradients obtained from the task classifier to perturb the input images~\cite{volpi2018generalizing,qiao2020learning,sinha2017certifying}. In doing so, the original data distribution is expanded, allowing the model to learn more generalizable features. Though this type of approach is often developed for tackling single-source DG, the idea can also be directly applied to multi-source scenarios.

\keypoint{Domain-Adversarial Gradients}
When it comes to multi-source DG where domain labels are provided, one can exploit domain-adversarial gradients to synthesize domain-agnostic images. For instance, \cite{shankar2018generalizing} trained a domain classifier and used its adversarial gradients to perturb the input images. Intuitively, by learning with domain-agnostic images, the task model is allowed to learn more domain-invariant patterns.

Since adversarial gradients-based perturbation is purposefully designed to be visually imperceptible~\cite{goodfellow2015explaining}, methods based on adversarial gradients are often criticized for not being able to simulate real-world domain shift, which is much more complicated than salt-and-pepper noise~\cite{zhou2020deep}. Furthermore, the computational cost is often doubled in these methods because the forward and backward passes need to be computed twice, which could pose serious efficiency issues for large neural networks. Below we discuss model-based methods that formulate the transformation $A(\cdot)$ using neural networks and can produce more diverse visual effects.

\keypoint{Random Augmentation Networks}
RandConv~\cite{xu2021robust} is based on the idea of using randomly initialized, single-layer convolutioinal neural network to transform the input images to ``novel domains.'' Since the weights are randomly sampled from a Gaussian distribution at each iteration and no learning is performed, the transformed images mainly contain random color distortions, which do not contain meaningful variations and are best to be mixed with the original images before passing to the task network.

\keypoint{Off-the-Shelf Style Transfer Models}
Taking advantage of the advances in style transfer~\cite{huang2017arbitrary}, several DG methods~\cite{somavarapu2020frustratingly,borlino2021rethinking,zhou2021stylematch} use off-the-shelf style transfer models like AdaIN~\cite{huang2017arbitrary} to represent $A(\cdot)$, which essentially maps images from one source domain to another for data augmentation. Instead of transferring image styles between source domains, one can exploit external styles to further diversify the source training data~\cite{yue2019domain}. Though these methods do not need to train the style transfer component, they still need domain labels for domain translation.

\keypoint{Learnable Augmentation Networks}
This group of methods aims to learn augmentation neural networks to synthesize new domains~\cite{zhou2020learning,zhou2020deep,carlucci2019hallucinating}. In~\cite{zhou2020deep,carlucci2019hallucinating}, domain-agnostic images are generated by maximizing the domain classification loss with respect to the image generator. In~\cite{zhou2020learning}, pseudo-novel domains are synthesized by maximizing for each source domain the domain distance measured by optimal transport~\cite{villani2008optimal} between the original and synthetic images.

\keypoint{Feature-Based Augmentation}
Though the above learnable augmentation models have shown promising results, their efficiency is a main concern as they need to train heavy image-to-image translation models. Another line of research focuses on feature-level augmentation~\cite{zhou2021mixstyle,zhou2021mixstylenn,mancini2020towards}. Motivated by the observation that image styles are captured in CNN feature statistics, MixStyle~\cite{zhou2021mixstyle,zhou2021mixstylenn} achieves style augmentation by mixing CNN feature statistics between instances of different domains. In~\cite{mancini2020towards}, Mixup~\cite{zhang2018mixup} is applied to mixing instances of different domains in both pixel and feature space.

\subsection{Ensemble Learning}\label{sec:methods;subsec:ensemble}
As an extensively studied topic in machine learning research, ensemble learning~\cite{zhou2012ensemble} typically learns multiple copies of the same model with different initialization weights or using different splits of training data, and uses their ensemble for prediction. Such a simple technique has been shown very effective in boosting the performance of a single model across a wide range of applications~\cite{krizhevsky2012imagenet,he2016deep,szegedy2015going}.

\keypoint{Exemplar-SVMs}
are a collection of SVM classifiers, each learned using one positive instance and all negative instances~\cite{malisiewicz2011ensemble}. As the ensemble of such exemplar SVMs have shown excellent generalization performance on the object detection task in~\cite{malisiewicz2011ensemble}, Xu et al.~\cite{xu2014exploiting} have extended exemplar-SVMs to DG. In particular, given a test sample the top-K exemplar classifiers that give the highest prediction scores (hence more confident) are selected for ensemble prediction. Such an idea of learning exemplar classifiers was also investigated in~\cite{niu2015multiview,niu2015visual} for DG.

\keypoint{Domain-Specific Neural Networks}
Since CNNs excel at discriminative feature learning~\cite{krizhevsky2012imagenet}, it is natural to replace hand-engineered SVM classifiers with CNN-based models for ensemble learning. A common practice is to learn domain-specific neural networks, each specializing in a source domain~\cite{ding2017deep,zhou2020domain}. Rather than learning an independent CNN for each source domain~\cite{ding2017deep}, it is more efficient, and makes more sense as well, to share between source domains some shallow layers~\cite{zhou2020domain}, which capture generic features~\cite{yosinski2014transferable}. Another question is how to compute the prediction. One can simply use the ensemble prediction averaged over all individuals with equal weights (e.g.,~\cite{zhou2020domain,d2018domain}). Alternatively, weighted averaging can be adopted where the weights are estimated by, for example, a source domain classifier aiming to measure the similarity of the target sample to each source domain~\cite{wang2020dofe}. Also, the weights can be used to determine the most confident candidate whose output will serve for final prediction~\cite{mancini2018best}.

\keypoint{Domain-Specific Batch Normalization}
In batch normalization (BN)~\cite{ioffe2015batch}, the statistics are computed on-the-fly during training and their moving averages are stored in buffers for inference. Since the statistics typically vary in different source domains, one could argue that mixing statistics of multiple source domains is detrimental to learning generalizable representations. One solution is to use domain-specific BNs~\cite{seo2020learning,mancini2018robust}, one for each source domain for collecting domain-specific statistics. This is equivalent to constructing domain-specific classifiers but with parameter sharing for most parts of a model except the normalization layers. Such a design was later adopted in~\cite{liu2020ms} for dealing with MRI segmentation. In~\cite{segu2020batch}, the domain-specific predictions are aggregated using as weights the distance between a test data's instance-level feature statistics and the source domain BN statistics.

\keypoint{Weight Averaging}
aggregates model weights at different time steps during training to form a single model at test time~\cite{izmailov2018averaging}. Unlike explicit ensemble learning where multiple models (or model parts) need to be trained, weight averaging is a more efficient solution as the model only needs to be trained once. In~\cite{cha2021domain}, the authors have demonstrated that weight averaging can greatly improve model robustness under domain shift. In fact, such a technique is orthogonal to many other DG approaches and can be applied as a post-processing method to further boost the DG performance.

\begin{figure}[t]
    \centering
    \includegraphics[width=\columnwidth]{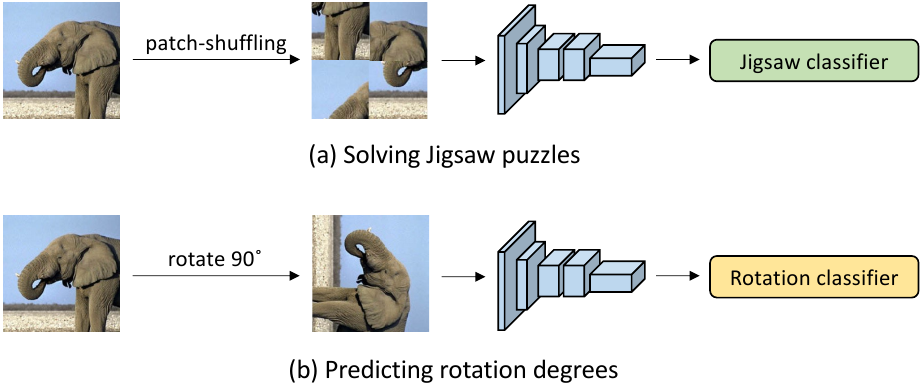}
    \caption{Common pretext tasks used for self-supervised learning in domain generalization. One can use a single pretext task, like solving Jigsaw puzzles~\cite{jigsaw_puzzles} or predicting rotations~\cite{gidaris2018unsupervised}, or combine multiple pretext tasks in a multi-task learning fashion.}
    \label{fig:pretext_tasks}
\end{figure}

\subsection{Self-Supervised Learning}\label{sec:methods;subsec:self_supervised}
Self-supervised learning is often referred to as learning with free labels generated from data itself (see~\cite{jing2020self} for a comprehensive survey on self-supervised learning). In computer vision, this can be achieved by teaching a model to predict the transformations applied to the image data, such as the shuffling order of patch-shuffled images~\cite{jigsaw_puzzles} or rotation degrees~\cite{gidaris2018unsupervised}. See Fig.~\ref{fig:pretext_tasks} for illustrations.

So why can self-supervised learning improve DG? An intuitive explanation is that solving pretext tasks allows a model to learn generic features regardless of the target task, and hence less over-fitting to domain-specific biases~\cite{cvpr19jigen}. An obvious advantage of self-supervised learning is that it can be applied to both single- and multi-source scenarios without requiring any domain labels.

\keypoint{Single Pretext Task}
In addition to using the standard classification loss, Carlucci et al.~\cite{cvpr19jigen} taught a neural network to solve the Jigsaw puzzles problem~\cite{jigsaw_puzzles}, hoping that the network can learn regularities that are more generalizable across domains. Similarly, Wang et al.~\cite{wang2020learning} used the Jigsaw solving task as an intrinsic supervision, together with an extrinsic supervision implemented using metric learning. Reconstruction has also been investigated for DG, such as learning an autoencoder to reconstruct image pixels/features~\cite{ghifary2015domain,maniyar2020zero}.

\keypoint{Multiple Pretext Tasks}
It is also possible to combine multiple pretext tasks. In~\cite{bucci2020self}, the authors combined two pretext tasks, namely solving Jigsaw puzzles and predicting rotations. In~\cite{albuquerque2020improving}, three pretext tasks are combined, namely reconstructing the Gabor filter's response, predicting rotations, and predicting feature cluster assignments~\cite{caron2018deep}. Overall, using multiple pretext tasks gives a better performance than using a single pretext task, as shown in~\cite{bucci2020self}.

Currently, these self-supervised learning-based DG methods have only been evaluated on the object recognition task. It is still unclear whether they will work on a wider range of OOD generalization tasks, which would be interesting to investigate in future work. Another concerns are that in general none of the existing pretext tasks is universal, and that the selection of pretext tasks is problem-specific. For instance, when the target domain shift is related to rotations, the model learned with the rotation prediction task will capture rotation-sensitive information, which is harmful to generalization.

Recent state-of-the-art self-supervised learning methods~\cite{he2020momentum,grill2020bootstrap} are mostly based on combining contrastive learning with data augmentation. The key idea is to pull together the same instance (image) undergone different transformations (e.g., random flip and color distortion) while push away different instances to learn instance-aware representations. Different from predicting transformations such as rotation, contrastive learning aims to learn transformation-invariant representations. Future work can explore whether invariances learned via contrastive learning can better adapt to OOD data.

\subsection{Learning Disentangled Representations}\label{sec:methods;subsec:disentangled}
Instead of forcing the entire model or features to be domain-invariant, which is challenging, one can relax this constraint by allowing some parts to be domain-specific, essentially learning disentangled representations. The existing approaches falling into this group are either based on decomposition~\cite{khosla2012undoing,li2017deeper,chattopadhyay2020learning,piratla2020efficient} or generative modeling~\cite{ilse2019diva,wang2020cross}, both requiring domain labels for feature disentanglement.

\keypoint{Decomposition}
An intuitive way to achieve disentangled representation learning is to decompose a model into two parts, one being domain-specific while the other being domain-agnostic. Based on SVMs, Khosla et al.~\cite{khosla2012undoing} decomposed a classifier into domain-specific biases and domain-agnostic weights, and only kept the latter when dealing with unseen domains. This approach was later extended to neural networks in~\cite{li2017deeper}. One can also design domain-specific modules such as in~\cite{chattopadhyay2020learning} where domain-specific binary masks are imposed on the final feature vector to distinguish between domain-specific and domain-invariant components. Another solution is to apply low-rank decomposition to a model's weight matrices in order to identify common features that are more generalizable~\cite{piratla2020efficient}.

\keypoint{Generative Modeling}
has been a powerful tool for learning disentangled representations~\cite{chen2016infogan}. In~\cite{ilse2019diva}, a variational autoencoder (VAE) is utilized to learn three independent latent subspaces for class, domain and object, respectively. In~\cite{wang2020cross}, two separate encoders are learned in an adversarial way to capture identity and domain information respectively for cross-domain face anti-spoofing.

\subsection{Regularization Strategies}\label{sec:methods;subsec:reg}
Some approaches focus on regularization strategies designed based on some heuristics. Wang et al.~\cite{wang2019learning} argued that generalizable features should capture the global structure/shape of objects rather than relying on local patches/textures, and therefore proposed to suppress the predictive power of auxiliary patch-wise CNNs (maximizing their classification errors), implemented as a stack of 1$\times$1 convolution layers. With a similar motivation, Huang et al.~\cite{huang2020self} iteratively masked out over-dominant features with large gradients, thus forcing the model to rely more on the remaining features. These methods do not require domain labels for learning, and are orthogonal to other DG methods like those based on domain alignment~\cite{motiian2017unified,shao2019multi} and data augmentation~\cite{volpi2019addressing,zhou2020learning,zhou2021mixstyle}. Therefore, one could potentially combine them to improve the performance in practice.

\subsection{Reinforcement Learning}\label{sec:methods;subsec:rl}
Domain shift in reinforcement learning (RL) not only occurs in visual appearance (color/style changes, etc.) but also in other aspects like dynamics (transition function) or rewards (e.g., gravity/friction changes)~\cite{kirk2021survey}. For visual domain shift, many of the DG methods surveyed above seem applicable for RL, such as data augmentation methods~\cite{zhou2021mixstyle,lee2019network}, but not for the latter that requires more problem-specific designs. Below we briefly discuss some representative generalization methods developed for RL. Please refer to \cite{kirk2021survey} for a more comprehensive survey.

\keypoint{Data Augmentation}
The main idea is to augment the visual signal sent to an RL agent to make it more domain-generalizable. A common approach is to use label-preserving transformations~\cite{laskin2020reinforcement,kostrikov2020image,hansen2021generalization} like color jittering or Cutout~\cite{devries2017cutout}. One can also implement the concept of domain randomization~\cite{tobin2017domain}, which visually randomizes an environment through, e.g., computer simulators~\cite{tobin2017domain} or random neural networks~\cite{lee2019network}. When convolutional neural networks are used, one can adopt the MixStyle~\cite{zhou2021mixstyle} approach discussed in Sec.~\ref{sec:methods;subsec:data_aug} to create ``novel'' domains in the feature space.

\keypoint{Self-Supervision}
Combining RL with self-supervised learning, which does not require manual labels, has also been explored. The general pipeline is to augment an RL model with auxiliary loss(es). For instance, Yarats et al.~\cite{yarats2021improving} proposed a reconstruction loss based on auto-encoders; Laskin et al.~\cite{laskin2020curl} combined RL with an unsupervised contrastive learning loss.

\section{Theories}\label{sec:theories}
Unlike domain adaptation in which plenty of learning bounds with theoretical guarantees have been proposed~\cite{redko2020survey}, bounding the risk for domain generalization (DG) is challenging due to the absence of target data. Nonetheless, some attempts have been made to address this problem, which are briefly reviewed in this section.

Most existing theoretical studies are subject to specific model classes, such as kernel methods~\cite{muandet2013domain,deshmukh2019generalization,blanchard2021domain,hu2020domain}, or have strong assumptions that cannot be simply applied to broader DG methods. In~\cite{li2020domain}, the latent feature space of all possible domains (including both source and target) is assumed to have a linear dependency, meaning that each domain is a linear combination of other domains. Based on the linear dependency assumption, a rank regularization method is proposed and combined with a distribution alignment method. In~\cite{albuquerque2019generalizing}, source domains are assumed to form a convex hull so that minimizing the maximum pairwise distance within the source domains would lead to a decrease in the distance between any two domains in the convex hull. In~\cite{rosenfeld2022online}, DG is cast into an online game where a player (model) minimizes the risk for a ``new'' distribution presented by an adversary at each time-step. In~\cite{vedantam2021empirical}, proxy measures that correlate well with ``true'' OOD generalization are investigated.

More recently, there are a couple of emerging studies~\cite{ye2021towards,li2022finding} that aim to provide more generic bounds, with more relaxed assumptions, for DG. In~\cite{ye2021towards}, feature distribution is quantified by two terms: \romannum{1}) a variation term measuring the stability of feature representations across domains; \romannum{2}) an informativeness term indicating the discriminativeness of feature representations (i.e., how well they can be used to distinguish different classes). Then, the error on unseen domains is bounded by an expansion function based on the variation term, subject to the learnability of feature representations measured using the informativeness term.

In~\cite{li2022finding}, the generalization gap is bounded in terms of the model's Rademacher complexity, suggesting that a lower model complexity with strong regularization can improve generalization in unseen domains---which echoes the findings in~\cite{gulrajani2020search}: properly regularized Empirical Risk Minimization with leave-one-domain-out cross-validation is a strong DG baseline.

\section{Future Research Directions}\label{sec:futdir}
So far we have covered the background on domain generalization (DG) in \S~\ref{sec:bg}---knowing what DG is about and how DG is typically evaluated under different settings/datasets---as well as gone though the existing methodologies developed over the last decade in \S~\ref{sec:methods}. The following questions would naturally arise: \romannum{1}) Has DG been solved? \romannum{2}) If not, how far are we from solving DG?

The answer is of course not---DG is a very challenging problem and is far from being solved. In this section, we aim to share some insights on future research directions, pointing out what have been missed in the current research and discussing what are worth exploring to further this field. Specifically, we discuss potential directions from three perspectives: \emph{model} (\S~\ref{sec:futdir;subsec:model}), \emph{learning} (\S~\ref{sec:futdir;subsec:learning}), and \emph{benchmarks} (\S~\ref{sec:futdir;subsec:benchmarks}).

\subsection{Model Architecture}\label{sec:futdir;subsec:model}

\keypoint{Dynamic Architectures}
The weights in a convolutional neural network (CNN), which serve as feature detectors, are normally fixed once learned from source domains. This may result in the representational power of a CNN model restricted to the seen domains while generalizing poorly when the image statistics in an unseen domain are significantly different. One potential solution is to develop \emph{dynamic} architectures~\cite{han2021dynamic}, e.g., with weights conditioned on the input~\cite{jia2016dynamic}. The key is to make neural networks' parameters (either partly or entirely) dependent on the input while ensuring that the model size is not too large to harm the efficiency. Dynamic architectures such as dynamic filter networks~\cite{jia2016dynamic} and conditional convolutions~\cite{yang2019condconv} have been shown effective on generic visual recognition tasks like classification and segmentation. It would be interesting to see whether such a flexible architecture can be used to cope with domain shift in DG.

\keypoint{Adaptive Normalization Layers}
Normalization layers~\cite{ioffe2015batch,ulyanov2016instance,ba2016layer} have been a core building block in contemporary neural networks. Following~\cite{luo2019differentiable}, a general formulation for different normalization layers can be written as $\gamma \frac{x-\mu}{\sigma} + \beta$, where $\mu$ and $\sigma$ denote mean and variance respectively; $\gamma$ and $\beta$ are learnable scaling and shift parameters respectively. Typically, $(\mu, \sigma)$ are computed on-the-fly during training but are saved in buffers using their moving averages for inference. Regardless of whether they are computed within each instance or based on a mini-batch, they can only represent the distribution of training data. The affine transformation parameters, i.e., $\gamma$ and $\beta$, are also learned for source data only. Therefore, a normalization layer's parameters are not guaranteed to work well under domain shift in unseen test data. It would be a promising direction to investigate how to make these parameters adaptive to unseen domains~\cite{park2019semantic}.

\subsection{Learning}\label{sec:futdir;subsec:learning}

\keypoint{Learning without Domain Labels}
Most existing methods leveraged domain labels in their models. However, in real-world applications it is possible that domain labels are difficult to obtain, e.g., web images crawled from the Internet are taken by arbitrary users with arbitrary domain characteristics and thus the domain labels are extremely difficult to define~\cite{niu2015visual}. In such scenarios where domain labels are missing, many top-performing DG approaches are not viable any more or the performance deteriorates~\cite{shi2021gradient}. Though this topic has been studied in the past (e.g.,~\cite{deecke2020latent,matsuura2020domain,zhou2021mixstyle}), methods that can deal with the absence of domain labels are still scarce and noncompetitive with methods that utilize domain labels. Considering that learning without domain labels is much more efficient and scalable, we encourage more future work to tackle this topic. We also suggest future work that uses domain labels evaluate the ability of functioning without proper domain labels---if applicable---like the random grouping experiment done in~\cite{shi2021gradient}.

\keypoint{Learning to Synthesize Novel Domains}
The DG performance can greatly benefit from increasing the diversity of source domains. This is also confirmed in a recent work~\cite{xu2021how} where the authors emphasized the importance of having diverse training distributions to out-of-distribution (OOD) generalization. However, in practice it is impossible to collect training data that cover all possible domains. As such, learning to synthesize novel domains can be a potential solution. Though this idea has been roughly explored in a couple of recent DG works~\cite{zhou2020learning,zhou2021mixstyle}, the results still have much room for improvements.

\keypoint{Avoiding Learning Shortcut}
Shortcut learning can be interpreted as a problem of learning ``easy'' representations that can perform well on training data but are irrelevant to the task~\cite{geirhos2020shortcut}. For example, given the task of distinguishing between digits blended with different colors, a neural network might be biased toward recognizing colors rather than the digit shapes during training, thus leading to poor generalization on unseen data~\cite{kim2019learning}. Such a problem can be intensified on multi-source data in DG as each source domain typically contains its own domain-specific bias. As a consequence, a DG model might simply learn to memorize the domain-specific biases, such as image styles~\cite{li2017deeper}, when tasked to differentiate between instances from different domains. The shortcut learning problem has been overlooked in DG.

\keypoint{Causal Representation Learning}
Currently, the common pipeline used in DG, as well as in many other fields, for representation learning is to learn a mapping $P(Y|X)$ by sampling data from the marginal distribution $P(X)$ with an objective to match the joint distribution $P(X,Y) = P(Y|X)P(X)$ (typically via maximum likelihood optimization). However, the learned representations have turned out to be lacking in the ability to adapt to OOD data~\cite{bengio2019meta}. A potential solution is to model the underlying causal variables (e.g., by autoencoder~\cite{bengio2019meta}) which cannot be directly observed but are much more stable and robust under distribution shift. This is closely related to the topic of causal representation learning, a recent trend in the machine learning community~\cite{scholkopf2021towards}.

\keypoint{Exploiting Side Information}
Side information (sometimes called meta-data) has been commonly used to boost the performance of a pattern recognition system. For example, depth information obtained from RGB-D sensors can be used alongside RGB images to improve the performance of, e.g., generic object detection~\cite{hoffman2016learning} or human detection~\cite{zhou2017detecting}. In DG, there exist a few works that utilize side information, such as attribute labels~\cite{gan2016learning} or object segmentation masks~\cite{zunino2020explainable}. In terms of attributes, they could be more generalizable because they capture mid- to low-level visual cues like colors, shapes and stripes, which are shared among different objects and less sensitive to domain biases~\cite{gan2016learning}. Notably, attributes have been widely used in zero-shot learning to recognize unseen classes~\cite{lampert2014attribute,xian2018zero}. In contrast, features learned for discrimination are usually too specific to objects, such as dog ears and human faces as found in top-layer CNN features~\cite{zeiler2014visualizing}, which are more likely to capture domain biases and hence less transferable between tasks~\cite{yosinski2014transferable}.

\keypoint{Transfer Learning}
A couple of recent works~\cite{chen2020automated,chen2021contrastive} have focused on the transfer learning perspective when designing DG methods for synthetic-to-real applications. Given a model pre-trained on large real datasets like ImageNet~\cite{deng2009imagenet}, the main goal is to learn new knowledge that is useful to the downstream task from synthetic data, and in the meantime, to maintain the knowledge on real images that was acquired from pre-training. Such a setting is closely related to learning-without-forgetting (LwF)~\cite{li2017learning}. In particular, a technique used in~\cite{chen2020automated} was borrowed from LwF~\cite{li2017learning}, i.e., minimizing the divergence between the new model's output and the old model's output to avoid erasing the pre-trained knowledge. Synthetic-to-real transfer learning is a realistic and practical setting but research in this direction has been less explored for DG.

\keypoint{Semi-Supervised Domain Generalization}
Most existing DG research assumes data collected from each source domain are fully annotated so the proposed methods are purely based on supervised learning, which are unable to cope with unlabeled data. However, in practice the size of labeled data could well be limited due to high annotation cost, but collecting abundant unlabeled data is much easier and cheaper. This leads to a more realistic and practical setting termed semi-supervised domain generalization~\cite{sharifi2020domain,zhou2021stylematch,zhou2021mixstylenn,liu2021semi,liu2022vmfnet}, which has recently picked up attention from the DG community. In~\cite{zhou2021stylematch}, pseudo-labels are assigned to unlabeled source data and an off-the-shelf style transfer model is used to augment the domain space. In~\cite{zhou2021mixstylenn}, feature statistics are mixed between labeled and pseudo-labeled source data for data augmentation. Since designing data-efficient, and yet generalizable learning systems is essential for practical applications, we believe semi-supervised domain generalizable is worth investigating for future work.

\keypoint{Open Domain Generalization}
is a recently introduced problem setting~\cite{shu2021open} where a model is learned from heterogeneous source domains with different label sets (with overlaps) and deployed in unseen domains for recognizing known classes while being able to reject unknown classes. This problem setting is related to existing heterogeneous DG~\cite{li2019episodic,zhou2020learning} but focuses on classification applications and emphasizes the ability to detect (reject) unknown classes, which is often studied in open-set recognition~\cite{geng2020recent}. In~\cite{shu2021open}, a variant of Mixup~\cite{zhang2018mixup} is proposed for data augmentation at both feature and label level, and a confidence threshold is used to reject test samples that likely belong to unknown classes.

\subsection{Benchmarks}\label{sec:futdir;subsec:benchmarks}

\keypoint{Incremental Learning + DG}
Most existing research on DG implicitly assumes that source domains are fixed and a model needs to be learned only once. However, in practice, it might well be the case that source domains are incrementally introduced, thus requiring incremental learning~\cite{wu2019large}. For instance, in cross-dataset person re-identification we might well have access to, say only two datasets at the beginning for model learning, e.g.,~Market1501~\cite{zheng2015scalable} and DukeMTMC-reID~\cite{ristani2016performance}, but later another dataset comes in, e.g.,~CUHK03~\cite{li2014deepreid}, which increases the number of source datasets from two to three. In this case, several problems need to be addressed, such as \romannum{1}) how to efficiently fine-tune the model on the new dataset without training from scratch using all available datasets, \romannum{2}) how to make sure the model does not over-fit the new dataset and forget the previously learned knowledge, and \romannum{3}) will the new dataset be beneficial or detrimental to the DG performance on the target domain.

\keypoint{Heterogeneous Domain Shift}
The current DG datasets mainly contain homogeneous domain shift, which means the source-source and source-target shifts are highly correlated with each other. For example, on PACS~\cite{li2017deeper} the source-source domain shift and the source-target domain shift are both related to image style changes; on Rotated MNIST~\cite{ghifary2015domain} rotation is the only cause of domain shift. However, in real-world scenarios the target domain shift is unpredictable and less likely to be correlated with the source domain shift, e.g., the source domains might be photo, art and sketch but the target domain might be images of novel viewpoints; or the source domains contain digit images with different rotations but the target domain images might be in a different font style or background. Such a setting, which we call heterogeneous domain shift, has never been brought up but is critical to practical applications.

\section{Conclusion}\label{sec:conclusion}
Domain generalization has been studied over a decade, with numerous methods developed in the literature across various application areas. Given the importance of domain generalization to the development of AI, it is imperative to make it clear that \romannum{1}) how this topic relates to neighboring fields like domain adaptation, \romannum{2}) how domain generalization is typically evaluated and benchmarked, and crucially, \romannum{3}) what the progress is in domain generalization. This timely and up-to-date survey answers these questions and we hope it can inspire future work to advance the field.

\keypoint{Acknowledgements}
This work is supported by NTU NAP, MOE AcRF Tier 2 (T2EP20221-0033), and under the RIE2020 Industry Alignment Fund – Industry Collaboration Projects (IAF-ICP) Funding Initiative, as well as cash and in-kind contribution from the industry partner(s). This work is also supported by National Natural Science Foundation of China (61876176, U1713208); the National Key Research and Development Program of China (No.~2020YFC2004800); Science and Technology Service Network Initiative of Chinese Academy of Sciences (KFJ-STS-QYZX-092); the Shanghai Committee of Science and Technology, China (Grant No.~20DZ1100800).


%





\ifCLASSOPTIONcaptionsoff
  \newpage
\fi


{
\bibliographystyle{IEEEtran}
\bibliography{reference}

\begin{thebibliography}{100}
\providecommand{\url}[1]{#1}
\csname url@samestyle\endcsname
\providecommand{\newblock}{\relax}
\providecommand{\bibinfo}[2]{#2}
\providecommand{\BIBentrySTDinterwordspacing}{\spaceskip=0pt\relax}
\providecommand{\BIBentryALTinterwordstretchfactor}{4}
\providecommand{\BIBentryALTinterwordspacing}{\spaceskip=\fontdimen2\font plus
\BIBentryALTinterwordstretchfactor\fontdimen3\font minus
  \fontdimen4\font\relax}
\providecommand{\BIBforeignlanguage}[2]{{%
\expandafter\ifx\csname l@#1\endcsname\relax
\typeout{** WARNING: IEEEtran.bst: No hyphenation pattern has been}%
\typeout{** loaded for the language `#1'. Using the pattern for}%
\typeout{** the default language instead.}%
\else
\language=\csname l@#1\endcsname
\fi
#2}}
\providecommand{\BIBdecl}{\relax}
\BIBdecl

\bibitem{moreno2012unifying}
J.~G. Moreno-Torres, T.~Raeder, R.~Alaiz-Rodr{\'\i}guez, N.~V. Chawla, and
  F.~Herrera, ``A unifying view on dataset shift in classification,''
  \emph{PR}, 2012.

\bibitem{recht2019imagenet}
B.~Recht, R.~Roelofs, L.~Schmidt, and V.~Shankar, ``Do imagenet classifiers
  generalize to imagenet?'' in \emph{ICML}, 2019.

\bibitem{ben2010theory}
S.~Ben-David, J.~Blitzer, K.~Crammer, A.~Kulesza, F.~Pereira, and J.~W.
  Vaughan, ``A theory of learning from different domains,'' \emph{ML}, 2010.

\bibitem{taori2020measuring}
R.~Taori, A.~Dave, V.~Shankar, N.~Carlini, B.~Recht, and L.~Schmidt,
  ``Measuring robustness to natural distribution shifts in image
  classification,'' in \emph{NeurIPS}, 2020.

\bibitem{blanchard2021domain}
G.~Blanchard, A.~A. Deshmukh, U.~Dogan, G.~Lee, and C.~Scott, ``Domain
  generalization by marginal transfer learning,'' \emph{JMLR}, 2021.

\bibitem{he2016deep}
K.~He, X.~Zhang, S.~Ren, and J.~Sun, ``Deep residual learning for image
  recognition,'' in \emph{CVPR}, 2016.

\bibitem{krizhevsky2012imagenet}
A.~Krizhevsky, I.~Sutskever, and G.~E. Hinton, ``Imagenet classification with
  deep convolutional neural networks,'' in \emph{NeurIPS}, 2012.

\bibitem{lecun2015deep}
Y.~LeCun, Y.~Bengio, and G.~Hinton, ``Deep learning,'' \emph{Nature}, 2015.

\bibitem{hendrycks2019benchmarking}
D.~Hendrycks and T.~Dietterich, ``Benchmarking neural network robustness to
  common corruptions and perturbations,'' in \emph{ICLR}, 2019.

\bibitem{yang2021generalized}
J.~Yang, K.~Zhou, Y.~Li, and Z.~Liu, ``Generalized out-of-distribution
  detection: A survey,'' \emph{arXiv preprint arXiv:2110.11334}, 2021.

\bibitem{deng2009imagenet}
J.~Deng, W.~Dong, R.~Socher, L.-J. Li, K.~Li, and L.~Fei-Fei, ``Imagenet: A
  large-scale hierarchical image database,'' in \emph{CVPR}, 2009.

\bibitem{saenko2010adapting}
K.~Saenko, B.~Kulis, M.~Fritz, and T.~Darrell, ``Adapting visual category
  models to new domains,'' in \emph{ECCV}, 2010.

\bibitem{lu2020stochastic}
Z.~Lu, Y.~Yang, X.~Zhu, C.~Liu, Y.-Z. Song, and T.~Xiang, ``Stochastic
  classifiers for unsupervised domain adaptation,'' in \emph{CVPR}, 2020.

\bibitem{saito2018maximum}
K.~Saito, K.~Watanabe, Y.~Ushiku, and T.~Harada, ``Maximum classifier
  discrepancy for unsupervised domain adaptation,'' in \emph{CVPR}, 2018.

\bibitem{ganin2015unsupervised}
Y.~Ganin and V.~S. Lempitsky, ``Unsupervised domain adaptation by
  backpropagation,'' in \emph{ICML}, 2015.

\bibitem{long2015learning}
M.~Long, Y.~Cao, J.~Wang, and M.~I. Jordan, ``Learning transferable features
  with deep adaptation networks,'' in \emph{ICML}, 2015.

\bibitem{liu2020open}
Z.~Liu, Z.~Miao, X.~Pan, X.~Zhan, D.~Lin, S.~X. Yu, and B.~Gong, ``Open
  compound domain adaptation,'' in \emph{CVPR}, 2020.

\bibitem{li2021learning}
B.~Li, Y.~Wang, S.~Zhang, D.~Li, T.~Darrell, K.~Keutzer, and H.~Zhao,
  ``Learning invariant representations and risks for semi-supervised domain
  adaptation,'' in \emph{CVPR}, 2021.

\bibitem{muandet2013domain}
K.~Muandet, D.~Balduzzi, and B.~Scholkopf, ``Domain generalization via
  invariant feature representation,'' in \emph{ICML}, 2013.

\bibitem{yue2019domain}
X.~Yue, Y.~Zhang, S.~Zhao, A.~Sangiovanni-Vincentelli, K.~Keutzer, and B.~Gong,
  ``Domain randomization and pyramid consistency: Simulation-to-real
  generalization without accessing target domain data,'' in \emph{ICCV}, 2019.

\bibitem{volpi2021continual}
R.~Volpi, D.~Larlus, and G.~Rogez, ``Continual adaptation of visual
  representations via domain randomization and meta-learning,'' in \emph{CVPR},
  2021.

\bibitem{blanchard2011generalizing}
G.~Blanchard, G.~Lee, and C.~Scott, ``Generalizing from several related
  classification tasks to a new unlabeled sample,'' in \emph{NeurIPS}, 2011.

\bibitem{zhou2021mixstyle}
K.~Zhou, Y.~Yang, Y.~Qiao, and T.~Xiang, ``Domain generalization with
  mixstyle,'' in \emph{ICLR}, 2021.

\bibitem{fan2021adversarially}
X.~Fan, Q.~Wang, J.~Ke, F.~Yang, B.~Gong, and M.~Zhou, ``Adversarially adaptive
  normalization for single domain generalization,'' in \emph{CVPR}, 2021.

\bibitem{zhang2021deep}
X.~Zhang, P.~Cui, R.~Xu, L.~Zhou, Y.~He, and Z.~Shen, ``Deep stable learning
  for out-of-distribution generalization,'' in \emph{CVPR}, 2021.

\bibitem{pandey2021domain}
P.~Pandey, M.~Raman, S.~Varambally, and P.~AP, ``Domain generalization via
  inference-time label-preserving target projections,'' in \emph{CVPR}, 2021.

\bibitem{shu2021open}
Y.~Shu, Z.~Cao, C.~Wang, J.~Wang, and M.~Long, ``Open domain generalization
  with domain-augmented meta-learning,'' in \emph{CVPR}, 2021.

\bibitem{zhou2021mixstylenn}
K.~Zhou, Y.~Yang, Y.~Qiao, and T.~Xiang, ``Mixstyle neural networks for domain
  generalization and adaptation,'' \emph{arXiv:2107.02053}, 2021.

\bibitem{zhou2021stylematch}
K.~Zhou, C.~C. Loy, and Z.~Liu, ``Semi-supervised domain generalization with
  stochastic stylematch,'' \emph{arXiv preprint arXiv:2106.00592}, 2021.

\bibitem{cha2021swad}
J.~Cha, S.~Chun, K.~Lee, H.-C. Cho, S.~Park, Y.~Lee, and S.~Park, ``Swad:
  Domain generalization by seeking flat minima,'' \emph{NeurIPS}, 2021.

\bibitem{li2018mmdaae}
H.~Li, S.~Jialin~Pan, S.~Wang, and A.~C. Kot, ``Domain generalization with
  adversarial feature learning,'' in \emph{CVPR}, 2018.

\bibitem{li2018ciddg}
Y.~Li, X.~Tiana, M.~Gong, Y.~Liu, T.~Liu, K.~Zhang, and D.~Tao, ``Deep domain
  generalization via conditional invariant adversarial networks,'' in
  \emph{ECCV}, 2018.

\bibitem{li2018learning}
D.~Li, Y.~Yang, Y.-Z. Song, and T.~M. Hospedales, ``Learning to generalize:
  Meta-learning for domain generalization,'' in \emph{AAAI}, 2018.

\bibitem{balaji2018metareg}
Y.~Balaji, S.~Sankaranarayanan, and R.~Chellappa, ``Metareg: Towards domain
  generalization using meta-regularization,'' in \emph{NeurIPS}, 2018.

\bibitem{zhou2020learning}
K.~Zhou, Y.~Yang, T.~Hospedales, and T.~Xiang, ``Learning to generate novel
  domains for domain generalization,'' in \emph{ECCV}, 2020.

\bibitem{zhou2020deep}
K.~Zhou, Y.~Yang, T.~M. Hospedales, and T.~Xiang, ``Deep domain-adversarial
  image generation for domain generalisation.'' in \emph{AAAI}, 2020.

\bibitem{li2017deeper}
D.~Li, Y.~Yang, Y.-Z. Song, and T.~M. Hospedales, ``Deeper, broader and artier
  domain generalization,'' in \emph{ICCV}, 2017.

\bibitem{feature_critic}
Y.~Li, Y.~Yang, W.~Zhou, and T.~Hospedales, ``Feature-critic networks for
  heterogeneous domain generalization,'' in \emph{ICML}, 2019.

\bibitem{volpi2019addressing}
R.~Volpi and V.~Murino, ``Addressing model vulnerability to distributional
  shifts over image transformation sets,'' in \emph{ICCV}, 2019.

\bibitem{shankar2018generalizing}
S.~Shankar, V.~Piratla, S.~Chakrabarti, S.~Chaudhuri, P.~Jyothi, and
  S.~Sarawagi, ``Generalizing across domains via cross-gradient training,'' in
  \emph{ICLR}, 2018.

\bibitem{liu2020ms}
Q.~Liu, Q.~Dou, L.~Yu, and P.~A. Heng, ``Ms-net: Multi-site network for
  improving prostate segmentation with heterogeneous mri data,'' \emph{TMI},
  2020.

\bibitem{liu2020shape}
Q.~Liu, Q.~Dou, and P.-A. Heng, ``Shape-aware meta-learning for generalizing
  prostate mri segmentation to unseen domains,'' in \emph{MICCAI}, 2020.

\bibitem{torralba2011unbiased}
A.~Torralba and A.~A. Efros, ``Unbiased look at dataset bias,'' in \emph{CVPR},
  2011.

\bibitem{fei2004learning}
L.~Fei-Fei, R.~Fergus, and P.~Perona, ``Learning generative visual models from
  few training examples: An incremental bayesian approach tested on 101 object
  categories,'' in \emph{CVPR-W}, 2004.

\bibitem{russell2008labelme}
B.~C. Russell, A.~Torralba, K.~P. Murphy, and W.~T. Freeman, ``Labelme: a
  database and web-based tool for image annotation,'' \emph{IJCV}, 2008.

\bibitem{khosla2012undoing}
A.~Khosla, T.~Zhou, T.~Malisiewicz, A.~Efros, and A.~Torralba, ``Undoing the
  damage of dataset bias,'' in \emph{ECCV}, 2012.

\bibitem{hendrycks2020augmix}
D.~Hendrycks, N.~Mu, E.~D. Cubuk, B.~Zoph, J.~Gilmer, and B.~Lakshminarayanan,
  ``Augmix: A simple data processing method to improve robustness and
  uncertainty,'' in \emph{ICLR}, 2020.

\bibitem{tang2021selfnorm}
Z.~Tang, Y.~Gao, Y.~Zhu, Z.~Zhang, M.~Li, and D.~Metaxas, ``Selfnorm and
  crossnorm for out-of-distribution robustness,'' \emph{arXiv preprint
  arXiv:2102.02811}, 2021.

\bibitem{cvpr19jigen}
F.~M. Carlucci, A.~D'Innocente, S.~Bucci, B.~Caputo, and T.~Tommasi, ``Domain
  generalization by solving jigsaw puzzles,'' in \emph{CVPR}, 2019.

\bibitem{bucci2020self}
S.~Bucci, A.~D'Innocente, Y.~Liao, F.~M. Carlucci, B.~Caputo, and T.~Tommasi,
  ``Self-supervised learning across domains,'' \emph{arXiv preprint
  arXiv:2007.12368}, 2020.

\bibitem{wang2019learning}
H.~Wang, Z.~He, Z.~C. Lipton, and E.~P. Xing, ``Learning robust representations
  by projecting superficial statistics out,'' in \emph{ICLR}, 2019.

\bibitem{huang2020self}
Z.~Huang, H.~Wang, E.~P. Xing, and D.~Huang, ``Self-challenging improves
  cross-domain generalization,'' in \emph{ECCV}, 2020.

\bibitem{ghifary2015domain}
M.~Ghifary, W.~B. Kleijn, M.~Zhang, and D.~Balduzzi, ``Domain generalization
  for object recognition with multi-task autoencoders,'' in \emph{ICCV}, 2015.

\bibitem{lecun1998mnist}
Y.~LeCun, L.~Bottou, Y.~Bengio, and P.~Haffner, ``Gradient-based learning
  applied to document recognition,'' in \emph{IEEE}, 1998.

\bibitem{netzer2011svhn}
Y.~Netzer, T.~Wang, A.~Coates, A.~Bissacco, B.~Wu, and A.~Y. Ng, ``Reading
  digits in natural images with unsupervised feature learning,'' in
  \emph{NeurIPS-W}, 2011.

\bibitem{fang2013unbiased}
C.~Fang, Y.~Xu, and D.~N. Rockmore, ``Unbiased metric learning: On the
  utilization of multiple datasets and web images for softening bias,'' in
  \emph{ICCV}, 2013.

\bibitem{everingham2010pascal}
M.~Everingham, L.~Van~Gool, C.~K. Williams, J.~Winn, and A.~Zisserman, ``The
  pascal visual object classes (voc) challenge,'' \emph{IJCV}, 2010.

\bibitem{xiao2010sun}
J.~Xiao, J.~Hays, K.~A. Ehinger, A.~Oliva, and A.~Torralba, ``Sun database:
  Large-scale scene recognition from abbey to zoo,'' in \emph{CVPR}, 2010.

\bibitem{office_home}
H.~Venkateswara, J.~Eusebio, S.~Chakraborty, and S.~Panchanathan, ``Deep
  hashing network for unsupervised domain adaptation,'' in \emph{CVPR}, 2017.

\bibitem{iccv19domainnet}
X.~Peng, Q.~Bai, X.~Xia, Z.~Huang, K.~Saenko, and B.~Wang, ``Moment matching
  for multi-source domain adaptation,'' in \emph{ICCV}, 2019.

\bibitem{zhou2020domain}
K.~Zhou, Y.~Yang, Y.~Qiao, and T.~Xiang, ``Domain adaptive ensemble learning,''
  \emph{arXiv preprint arXiv:2003.07325}, 2020.

\bibitem{peng2017visda}
X.~Peng, B.~Usman, N.~Kaushik, J.~Hoffman, D.~Wang, and K.~Saenko, ``Visda: The
  visual domain adaptation challenge,'' \emph{arXiv preprint arXiv:1710.06924},
  2017.

\bibitem{hendrycks2021many}
D.~Hendrycks, S.~Basart, N.~Mu, S.~Kadavath, F.~Wang, E.~Dorundo, R.~Desai,
  T.~Zhu, S.~Parajuli, M.~Guo \emph{et~al.}, ``The many faces of robustness: A
  critical analysis of out-of-distribution generalization,'' in \emph{ICCV},
  2021.

\bibitem{hendrycks2021natural}
D.~Hendrycks, K.~Zhao, S.~Basart, J.~Steinhardt, and D.~Song, ``Natural
  adversarial examples,'' in \emph{CVPR}, 2021.

\bibitem{beery2018recognition}
S.~Beery, G.~Van~Horn, and P.~Perona, ``Recognition in terra incognita,'' in
  \emph{ECCV}, 2018.

\bibitem{zhang2022nico++}
X.~Zhang, L.~Zhou, R.~Xu, P.~Cui, Z.~Shen, and H.~Liu, ``Nico++: Towards better
  benchmarking for domain generalization,'' \emph{arXiv preprint
  arXiv:2204.08040}, 2022.

\bibitem{rebuffi2017learning}
S.-A. Rebuffi, H.~Bilen, and A.~Vedaldi, ``Learning multiple visual domains
  with residual adapters,'' in \emph{NeurIPS}, 2017.

\bibitem{weinland2006free}
D.~Weinland, R.~Ronfard, and E.~Boyer, ``Free viewpoint action recognition
  using motion history volumes,'' \emph{CVIU}, 2006.

\bibitem{soomro2012ucf101}
K.~Soomro, A.~R. Zamir, and M.~Shah, ``Ucf101: A dataset of 101 human actions
  classes from videos in the wild,'' \emph{arXiv preprint arXiv:1212.0402},
  2012.

\bibitem{kuehne2011hmdb}
H.~Kuehne, H.~Jhuang, E.~Garrote, T.~Poggio, and T.~Serre, ``Hmdb: a large
  video database for human motion recognition,'' in \emph{ICCV}, 2011.

\bibitem{yao2019adversarial}
Z.~Yao, Y.~Wang, X.~Du, M.~Long, and J.~Wang, ``Adversarial pyramid network for
  video domain generalization,'' \emph{arXiv preprint arXiv:1912.03716}, 2019.

\bibitem{ros2016synthia}
G.~Ros, L.~Sellart, J.~Materzynska, D.~Vazquez, and A.~M. Lopez, ``The synthia
  dataset: A large collection of synthetic images for semantic segmentation of
  urban scenes,'' in \emph{CVPR}, 2016.

\bibitem{volpi2018generalizing}
R.~Volpi, H.~Namkoong, O.~Sener, J.~Duchi, V.~Murino, and S.~Savarese,
  ``Generalizing to unseen domains via adversarial data augmentation,'' in
  \emph{NeurIPS}, 2018.

\bibitem{richter2016playing}
S.~R. Richter, V.~Vineet, S.~Roth, and V.~Koltun, ``Playing for data: Ground
  truth from computer games,'' in \emph{ECCV}, 2016.

\bibitem{cordts2016cityscapes}
M.~Cordts, M.~Omran, S.~Ramos, T.~Rehfeld, M.~Enzweiler, R.~Benenson,
  U.~Franke, S.~Roth, and B.~Schiele, ``The cityscapes dataset for semantic
  urban scene understanding,'' in \emph{CVPR}, 2016.

\bibitem{zheng2015scalable}
L.~Zheng, L.~Shen, L.~Tian, S.~Wang, J.~Wang, and Q.~Tian, ``Scalable person
  re-identification: A benchmark,'' in \emph{ICCV}, 2015.

\bibitem{ristani2016performance}
E.~Ristani, F.~Solera, R.~Zou, R.~Cucchiara, and C.~Tomasi, ``Performance
  measures and a data set for multi-target, multi-camera tracking,'' in
  \emph{ECCV}, 2016.

\bibitem{shi2020towards}
Y.~Shi, X.~Yu, K.~Sohn, M.~Chandraker, and A.~K. Jain, ``Towards universal
  representation learning for deep face recognition,'' in \emph{CVPR}, 2020.

\bibitem{zhang2012face}
Z.~Zhang, J.~Yan, S.~Liu, Z.~Lei, D.~Yi, and S.~Z. Li, ``A face antispoofing
  database with diverse attacks,'' in \emph{ICB}, 2012.

\bibitem{boulkenafet2017oulu}
Z.~Boulkenafet, J.~Komulainen, L.~Li, X.~Feng, and A.~Hadid, ``Oulu-npu: A
  mobile face presentation attack database with real-world variations,'' in
  \emph{FG}, 2017.

\bibitem{wen2015face}
D.~Wen, H.~Han, and A.~K. Jain, ``Face spoof detection with image distortion
  analysis,'' \emph{TIFS}, 2015.

\bibitem{chingovska2012effectiveness}
I.~Chingovska, A.~Anjos, and S.~Marcel, ``On the effectiveness of local binary
  patterns in face anti-spoofing,'' in \emph{BIOSIG}, 2012.

\bibitem{sainath2015convolutional}
T.~Sainath and C.~Parada, ``Convolutional neural networks for small-footprint
  keyword spotting,'' 2015.

\bibitem{blitzer2006domain}
J.~Blitzer, R.~McDonald, and F.~Pereira, ``Domain adaptation with structural
  correspondence learning,'' in \emph{EMNLP}, 2006.

\bibitem{wilds2020}
P.~W. Koh, S.~Sagawa, H.~Marklund, S.~M. Xie, M.~Zhang, A.~Balsubramani, W.~Hu,
  M.~Yasunaga, R.~L. Phillips, S.~Beery, J.~Leskovec, A.~Kundaje, E.~Pierson,
  S.~Levine, C.~Finn, and P.~Liang, ``Wilds: A benchmark of in-the-wild
  distribution shifts,'' \emph{arXiv preprint arXiv:2012.07421}, 2020.

\bibitem{bandi2018detection}
P.~Bandi, O.~Geessink, Q.~Manson, M.~Van~Dijk, M.~Balkenhol, M.~Hermsen, B.~E.
  Bejnordi, B.~Lee, K.~Paeng, A.~Zhong \emph{et~al.}, ``From detection of
  individual metastases to classification of lymph node status at the patient
  level: the camelyon17 challenge,'' \emph{TMI}, 2018.

\bibitem{christie2018functional}
G.~Christie, N.~Fendley, J.~Wilson, and R.~Mukherjee, ``Functional map of the
  world,'' in \emph{CVPR}, 2018.

\bibitem{beery2020iwildcam}
S.~Beery, E.~Cole, and A.~Gjoka, ``The iwildcam 2020 competition dataset,''
  \emph{arXiv preprint arXiv:2004.10340}, 2020.

\bibitem{mahajan2021domain}
D.~Mahajan, S.~Tople, and A.~Sharma, ``Domain generalization using causal
  matching,'' in \emph{ICML}, 2021.

\bibitem{wang2017chestx}
X.~Wang, Y.~Peng, L.~Lu, Z.~Lu, M.~Bagheri, and R.~M. Summers, ``Chestx-ray8:
  Hospital-scale chest x-ray database and benchmarks on weakly-supervised
  classification and localization of common thorax diseases,'' in \emph{CVPR},
  2017.

\bibitem{irvin2019chexpert}
J.~Irvin, P.~Rajpurkar, M.~Ko, Y.~Yu, S.~Ciurea-Ilcus, C.~Chute, H.~Marklund,
  B.~Haghgoo, R.~Ball, K.~Shpanskaya \emph{et~al.}, ``Chexpert: A large chest
  radiograph dataset with uncertainty labels and expert comparison,'' in
  \emph{AAAI}, 2019.

\bibitem{cobbe2019quantifying}
K.~Cobbe, O.~Klimov, C.~Hesse, T.~Kim, and J.~Schulman, ``Quantifying
  generalization in reinforcement learning,'' in \emph{ICML}, 2019.

\bibitem{cobbe2020leveraging}
K.~Cobbe, C.~Hesse, J.~Hilton, and J.~Schulman, ``Leveraging procedural
  generation to benchmark reinforcement learning,'' in \emph{ICML}, 2020.

\bibitem{li2019episodic}
D.~Li, J.~Zhang, Y.~Yang, C.~Liu, Y.-Z. Song, and T.~M. Hospedales, ``Episodic
  training for domain generalization,'' in \emph{ICCV}, 2019.

\bibitem{hoffman2018cycada}
J.~Hoffman, E.~Tzeng, T.~Park, J.-Y. Zhu, P.~Isola, K.~Saenko, A.~Efros, and
  T.~Darrell, ``Cycada: Cycle-consistent adversarial domain adaptation,'' in
  \emph{ICML}, 2018.

\bibitem{gong2018dlow}
R.~Gong, W.~Li, Y.~Chen, and L.~Van~Gool, ``Dlow: Domain flow for adaptation
  and generalization,'' in \emph{CVPR}, 2019.

\bibitem{sun2019learning}
Y.~Sun, L.~Zheng, Y.~Li, Y.~Yang, Q.~Tian, and S.~Wang, ``Learning part-based
  convolutional features for person re-identification,'' \emph{TPAMI}, 2019.

\bibitem{li2019scalable}
W.~Li, X.~Zhu, and S.~Gong, ``Scalable person re-identification by harmonious
  attention,'' \emph{IJCV}, 2019.

\bibitem{chen2019abd}
T.~Chen, S.~Ding, J.~Xie, Y.~Yuan, W.~Chen, Y.~Yang, Z.~Ren, and Z.~Wang,
  ``Abd-net: Attentive but diverse person re-identification,'' in \emph{ICCV},
  2019.

\bibitem{zhou2019osnet}
K.~Zhou, Y.~Yang, A.~Cavallaro, and T.~Xiang, ``Omni-scale feature learning for
  person re-identification,'' in \emph{ICCV}, 2019.

\bibitem{chang2018multi}
X.~Chang, T.~M. Hospedales, and T.~Xiang, ``Multi-level factorisation net for
  person re-identification,'' in \emph{CVPR}, 2018.

\bibitem{zhou2021osnet}
K.~Zhou, Y.~Yang, A.~Cavallaro, and T.~Xiang, ``Learning generalisable
  omni-scale representations for person re-identification,'' \emph{TPAMI},
  2021.

\bibitem{zhao2021learning}
Y.~Zhao, Z.~Zhong, F.~Yang, Z.~Luo, Y.~Lin, S.~Li, and N.~Sebe, ``Learning to
  generalize unseen domains via memory-based multi-source meta-learning for
  person re-identification,'' in \emph{CVPR}, 2021.

\bibitem{choi2020meta}
S.~Choi, T.~Kim, M.~Jeong, H.~Park, and C.~Kim, ``Meta batch-instance
  normalization for generalizable person re-identification,'' \emph{arXiv
  preprint arXiv:2011.14670}, 2020.

\bibitem{taigman2014deepface}
Y.~Taigman, M.~Yang, M.~Ranzato, and L.~Wolf, ``Deepface: Closing the gap to
  human-level performance in face verification,'' in \emph{CVPR}, 2014.

\bibitem{sun2014deep}
Y.~Sun, X.~Wang, and X.~Tang, ``Deep learning face representation from
  predicting 10,000 classes,'' in \emph{CVPR}, 2014.

\bibitem{wen2016discriminative}
Y.~Wen, K.~Zhang, Z.~Li, and Y.~Qiao, ``A discriminative feature learning
  approach for deep face recognition,'' in \emph{ECCV}, 2016.

\bibitem{guo2016ms}
Y.~Guo, L.~Zhang, Y.~Hu, X.~He, and J.~Gao, ``Ms-celeb-1m: A dataset and
  benchmark for large-scale face recognition,'' in \emph{ECCV}, 2016.

\bibitem{wolf2011face}
L.~Wolf, T.~Hassner, and I.~Maoz, ``Face recognition in unconstrained videos
  with matched background similarity,'' in \emph{CVPR}, 2011.

\bibitem{kalka2018ijb}
N.~D. Kalka, B.~Maze, J.~A. Duncan, K.~O’Connor, S.~Elliott, K.~Hebert,
  J.~Bryan, and A.~K. Jain, ``Ijb--s: Iarpa janus surveillance video
  benchmark,'' in \emph{BTAS}, 2018.

\bibitem{cheng2018low}
Z.~Cheng, X.~Zhu, and S.~Gong, ``Low-resolution face recognition,'' in
  \emph{ACCV}, 2018.

\bibitem{klare2015pushing}
B.~F. Klare, B.~Klein, E.~Taborsky, A.~Blanton, J.~Cheney, K.~Allen,
  P.~Grother, A.~Mah, and A.~K. Jain, ``Pushing the frontiers of unconstrained
  face detection and recognition: Iarpa janus benchmark a,'' in \emph{CVPR},
  2015.

\bibitem{maze2018iarpa}
B.~Maze, J.~Adams, J.~A. Duncan, N.~Kalka, T.~Miller, C.~Otto, A.~K. Jain,
  W.~T. Niggel, J.~Anderson, J.~Cheney \emph{et~al.}, ``Iarpa janus
  benchmark-c: Face dataset and protocol,'' in \emph{ICB}, 2018.

\bibitem{sengupta2016frontal}
S.~Sengupta, J.-C. Chen, C.~Castillo, V.~M. Patel, R.~Chellappa, and D.~W.
  Jacobs, ``Frontal to profile face verification in the wild,'' in \emph{WACV},
  2016.

\bibitem{kemelmacher2016megaface}
I.~Kemelmacher-Shlizerman, S.~M. Seitz, D.~Miller, and E.~Brossard, ``The
  megaface benchmark: 1 million faces for recognition at scale,'' in
  \emph{CVPR}, 2016.

\bibitem{yang2014learn}
J.~Yang, Z.~Lei, and S.~Z. Li, ``Learn convolutional neural network for face
  anti-spoofing,'' \emph{arXiv preprint arXiv:1408.5601}, 2014.

\bibitem{shao2019multi}
R.~Shao, X.~Lan, J.~Li, and P.~C. Yuen, ``Multi-adversarial discriminative deep
  domain generalization for face presentation attack detection,'' in
  \emph{CVPR}, 2019.

\bibitem{shi2021gradient}
Y.~Shi, J.~Seely, P.~H. Torr, N.~Siddharth, A.~Hannun, N.~Usunier, and
  G.~Synnaeve, ``Gradient matching for domain generalization,'' \emph{arXiv
  preprint arXiv:2104.09937}, 2021.

\bibitem{robey2021model}
A.~Robey, G.~J. Pappas, and H.~Hassani, ``Model-based domain generalization,''
  \emph{NeurIPS}, 2021.

\bibitem{yao2022improving}
H.~Yao, Y.~Wang, S.~Li, L.~Zhang, W.~Liang, J.~Zou, and C.~Finn, ``Improving
  out-of-distribution robustness via selective augmentation,'' \emph{arXiv
  preprint arXiv:2201.00299}, 2022.

\bibitem{dou2019domain}
Q.~Dou, D.~C. Castro, K.~Kamnitsas, and B.~Glocker, ``Domain generalization via
  model-agnostic learning of semantic features,'' in \emph{NeurIPS}, 2019.

\bibitem{mazoure2021improving}
B.~Mazoure, I.~Kostrikov, O.~Nachum, and J.~Tompson, ``Improving zero-shot
  generalization in offline reinforcement learning using generalized similarity
  functions,'' \emph{arXiv preprint arXiv:2111.14629}, 2021.

\bibitem{hansen2021generalization}
N.~Hansen and X.~Wang, ``Generalization in reinforcement learning by soft data
  augmentation,'' in \emph{ICRA}, 2021.

\bibitem{deitke2020robothor}
M.~Deitke, W.~Han, A.~Herrasti, A.~Kembhavi, E.~Kolve, R.~Mottaghi,
  J.~Salvador, D.~Schwenk, E.~VanderBilt, M.~Wallingford \emph{et~al.},
  ``Robothor: An open simulation-to-real embodied ai platform,'' in
  \emph{CVPR}, 2020.

\bibitem{kirk2021survey}
R.~Kirk, A.~Zhang, E.~Grefenstette, and T.~Rockt{\"a}schel, ``A survey of
  generalisation in deep reinforcement learning,'' \emph{arXiv preprint
  arXiv:2111.09794}, 2021.

\bibitem{sagawa2019distributionally}
S.~Sagawa, P.~W. Koh, T.~B. Hashimoto, and P.~Liang, ``Distributionally robust
  neural networks for group shifts: On the importance of regularization for
  worst-case generalization,'' \emph{arXiv preprint arXiv:1911.08731}, 2019.

\bibitem{krueger2021out}
D.~Krueger, E.~Caballero, J.-H. Jacobsen, A.~Zhang, J.~Binas, D.~Zhang,
  R.~Le~Priol, and A.~Courville, ``Out-of-distribution generalization via risk
  extrapolation (rex),'' in \emph{ICML}, 2021.

\bibitem{gulrajani2020search}
I.~Gulrajani and D.~Lopez-Paz, ``In search of lost domain generalization,''
  \emph{arXiv preprint arXiv:2007.01434}, 2020.

\bibitem{yang2017deep}
Y.~Yang and T.~Hospedales, ``Deep multi-task representation learning: A tensor
  factorisation approach,'' in \emph{ICLR}, 2017.

\bibitem{mallya2018piggyback}
A.~Mallya, D.~Davis, and S.~Lazebnik, ``Piggyback: Adapting a single network to
  multiple tasks by learning to mask weights,'' in \emph{ECCV}, 2018.

\bibitem{liu2019end}
S.~Liu, E.~Johns, and A.~J. Davison, ``End-to-end multi-task learning with
  attention,'' in \emph{CVPR}, 2019.

\bibitem{guo2020learning}
P.~Guo, C.-Y. Lee, and D.~Ulbricht, ``Learning to branch for multi-task
  learning,'' in \emph{ICML}, 2020.

\bibitem{sun2020adashare}
X.~Sun, R.~Panda, R.~Feris, and K.~Saenko, ``Adashare: Learning what to share
  for efficient deep multi-task learning,'' in \emph{NeurIPS}, 2020.

\bibitem{wang2020learning}
S.~Wang, L.~Yu, C.~Li, C.-W. Fu, and P.-A. Heng, ``Learning from extrinsic and
  intrinsic supervisions for domain generalization,'' in \emph{ECCV}, 2020.

\bibitem{albuquerque2020improving}
I.~Albuquerque, N.~Naik, J.~Li, N.~Keskar, and R.~Socher, ``Improving
  out-of-distribution generalization via multi-task self-supervised
  pretraining,'' \emph{arXiv preprint arXiv:2003.13525}, 2020.

\bibitem{pan2009survey}
S.~J. Pan and Q.~Yang, ``A survey on transfer learning,'' \emph{TKDE}, 2009.

\bibitem{zhu2015aligning}
Y.~Zhu, R.~Kiros, R.~Zemel, R.~Salakhutdinov, R.~Urtasun, A.~Torralba, and
  S.~Fidler, ``Aligning books and movies: Towards story-like visual
  explanations by watching movies and reading books,'' in \emph{ICCV}, 2015.

\bibitem{girshick2014rich}
R.~Girshick, J.~Donahue, T.~Darrell, and J.~Malik, ``Rich feature hierarchies
  for accurate object detection and semantic segmentation,'' in \emph{CVPR},
  2014.

\bibitem{yosinski2014transferable}
J.~Yosinski, J.~Clune, Y.~Bengio, and H.~Lipson, ``How transferable are
  features in deep neural networks?'' in \emph{NeurIPS}, 2014.

\bibitem{donahue2014decaf}
J.~Donahue, Y.~Jia, O.~Vinyals, J.~Hoffman, N.~Zhang, E.~Tzeng, and T.~Darrell,
  ``Decaf: A deep convolutional activation feature for generic visual
  recognition,'' in \emph{ICML}, 2014.

\bibitem{chen2020automated}
W.~Chen, Z.~Yu, Z.~Wang, and A.~Anandkumar, ``Automated synthetic-to-real
  generalization,'' in \emph{ICML}, 2020.

\bibitem{chen2021contrastive}
W.~Chen, Z.~Yu, S.~D. Mello, S.~Liu, J.~M. Alvarez, Z.~Wang, and A.~Anandkumar,
  ``Contrastive syn-to-real generalization,'' in \emph{ICLR}, 2021.

\bibitem{lampert2014attribute}
C.~H. Lampert, H.~Nickisch, and S.~Harmeling, ``Attribute-based classification
  for zero-shot visual object categorization,'' \emph{TPAMI}, 2014.

\bibitem{zhou2021coop}
K.~Zhou, J.~Yang, C.~C. Loy, and Z.~Liu, ``Learning to prompt for
  vision-language models,'' \emph{arXiv preprint arXiv:2109.01134}, 2021.

\bibitem{zhou2022cocoop}
------, ``Conditional prompt learning for vision-language models,'' in
  \emph{CVPR}, 2022.

\bibitem{chao2016empirical}
W.-L. Chao, S.~Changpinyo, B.~Gong, and F.~Sha, ``An empirical study and
  analysis of generalized zero-shot learning for object recognition in the
  wild,'' in \emph{ECCV}, 2016.

\bibitem{mancini2020towards}
M.~Mancini, Z.~Akata, E.~Ricci, and B.~Caputo, ``Towards recognizing unseen
  categories in unseen domains,'' in \emph{ECCV}, 2020.

\bibitem{xian2018zero}
Y.~Xian, C.~H. Lampert, B.~Schiele, and Z.~Akata, ``Zero-shot learning—a
  comprehensive evaluation of the good, the bad and the ugly,'' \emph{TPAMI},
  2018.

\bibitem{gan2016learning}
C.~Gan, T.~Yang, and B.~Gong, ``Learning attributes equals multi-source domain
  generalization,'' in \emph{CVPR}, 2016.

\bibitem{gong2012geodesic}
B.~Gong, Y.~Shi, F.~Sha, and K.~Grauman, ``Geodesic flow kernel for
  unsupervised domain adaptation,'' in \emph{CVPR}, 2012.

\bibitem{long2016unsupervised}
M.~Long, H.~Zhu, J.~Wang, and M.~I. Jordan, ``Unsupervised domain adaptation
  with residual transfer networks,'' in \emph{NeurIPS}, 2016.

\bibitem{balaji2019normalized}
Y.~Balaji, R.~Chellappa, and S.~Feizi, ``Normalized wasserstein for mixture
  distributions with applications in adversarial learning and domain
  adaptation,'' in \emph{ICCV}, 2019.

\bibitem{kang2019contrastive}
G.~Kang, L.~Jiang, Y.~Yang, and A.~G. Hauptmann, ``Contrastive adaptation
  network for unsupervised domain adaptation,'' in \emph{CVPR}, 2019.

\bibitem{kulis2011you}
B.~Kulis, K.~Saenko, and T.~Darrell, ``What you saw is not what you get: Domain
  adaptation using asymmetric kernel transforms,'' in \emph{CVPR}, 2011.

\bibitem{peng2018zero}
K.-C. Peng, Z.~Wu, and J.~Ernst, ``Zero-shot deep domain adaptation,'' in
  \emph{ECCV}, 2018.

\bibitem{panareda2017open}
P.~Panareda~Busto and J.~Gall, ``Open set domain adaptation,'' in \emph{ICCV},
  2017.

\bibitem{cao2018partial}
Z.~Cao, L.~Ma, M.~Long, and J.~Wang, ``Partial adversarial domain adaptation,''
  in \emph{ECCV}, 2018.

\bibitem{you2019universal}
K.~You, M.~Long, Z.~Cao, J.~Wang, and M.~I. Jordan, ``Universal domain
  adaptation,'' in \emph{CVPR}, 2019.

\bibitem{wang2020tent}
D.~Wang, E.~Shelhamer, S.~Liu, B.~Olshausen, and T.~Darrell, ``Tent: Fully
  test-time adaptation by entropy minimization,'' \emph{arXiv preprint
  arXiv:2006.10726}, 2020.

\bibitem{kundu2020universal}
J.~N. Kundu, N.~Venkat, R.~V. Babu \emph{et~al.}, ``Universal source-free
  domain adaptation,'' in \emph{CVPR}, 2020.

\bibitem{zhang2021memo}
M.~Zhang, S.~Levine, and C.~Finn, ``Memo: Test time robustness via adaptation
  and augmentation,'' \emph{arXiv preprint arXiv:2110.09506}, 2021.

\bibitem{iwasawa2021test}
Y.~Iwasawa and Y.~Matsuo, ``Test-time classifier adjustment module for
  model-agnostic domain generalization,'' \emph{NeurIPS}, 2021.

\bibitem{sun2020test}
Y.~Sun, X.~Wang, Z.~Liu, J.~Miller, A.~Efros, and M.~Hardt, ``Test-time
  training with self-supervision for generalization under distribution
  shifts,'' in \emph{ICML}, 2020.

\bibitem{erfani2016robust}
S.~Erfani, M.~Baktashmotlagh, M.~Moshtaghi, X.~Nguyen, C.~Leckie, J.~Bailey,
  and R.~Kotagiri, ``Robust domain generalisation by enforcing distribution
  invariance,'' in \emph{IJCAI}, 2016.

\bibitem{ghifary2017scatter}
M.~Ghifary, D.~Balduzzi, W.~B. Kleijn, and M.~Zhang, ``Scatter component
  analysis: A unified framework for domain adaptation and domain
  generalization,'' \emph{TPAMI}, 2017.

\bibitem{li2018domain}
Y.~Li, M.~Gong, X.~Tian, T.~Liu, and D.~Tao, ``Domain generalization via
  conditional invariant representations,'' in \emph{AAAI}, 2018.

\bibitem{jin2020feature}
X.~Jin, C.~Lan, W.~Zeng, and Z.~Chen, ``Feature alignment and restoration for
  domain generalization and adaptation,'' \emph{arXiv preprint
  arXiv:2006.12009}, 2020.

\bibitem{hu2020domain}
S.~Hu, K.~Zhang, Z.~Chen, and L.~Chan, ``Domain generalization via multidomain
  discriminant analysis,'' in \emph{UAI}, 2020.

\bibitem{motiian2017unified}
S.~Motiian, M.~Piccirilli, D.~A. Adjeroh, and G.~Doretto, ``Unified deep
  supervised domain adaptation and generalization,'' in \emph{ICCV}, 2017.

\bibitem{yoon2019generalizable}
C.~Yoon, G.~Hamarneh, and R.~Garbi, ``Generalizable feature learning in the
  presence of data bias and domain class imbalance with application to skin
  lesion classification,'' in \emph{MICCAI}, 2019.

\bibitem{mahajan2020domain}
D.~Mahajan, S.~Tople, and A.~Sharma, ``Domain generalization using causal
  matching,'' \emph{arXiv preprint arXiv:2006.07500}, 2020.

\bibitem{wang2020respecting}
Z.~Wang, M.~Loog, and J.~van Gemert, ``Respecting domain relations: Hypothesis
  invariance for domain generalization,'' \emph{arXiv preprint
  arXiv:2010.07591}, 2020.

\bibitem{li2020domain}
H.~Li, Y.~Wang, R.~Wan, S.~Wang, T.-Q. Li, and A.~C. Kot, ``Domain
  generalization for medical imaging classification with linear-dependency
  regularization,'' in \emph{NeurIPS}, 2020.

\bibitem{rahman2020correlation}
M.~M. Rahman, C.~Fookes, M.~Baktashmotlagh, and S.~Sridharan,
  ``Correlation-aware adversarial domain adaptation and generalization,''
  \emph{PR}, 2020.

\bibitem{albuquerque2019generalizing}
I.~Albuquerque, J.~Monteiro, M.~Darvishi, T.~H. Falk, and I.~Mitliagkas,
  ``Generalizing to unseen domains via distribution matching,'' \emph{arXiv
  preprint arXiv:1911.00804}, 2019.

\bibitem{deng2020representation}
Z.~Deng, F.~Ding, C.~Dwork, R.~Hong, G.~Parmigiani, P.~Patil, and P.~Sur,
  ``Representation via representations: Domain generalization via adversarially
  learned invariant representations,'' \emph{arXiv preprint arXiv:2006.11478},
  2020.

\bibitem{matsuura2020domain}
T.~Matsuura and T.~Harada, ``Domain generalization using a mixture of multiple
  latent domains,'' in \emph{AAAI}, 2020.

\bibitem{jia2020single}
Y.~Jia, J.~Zhang, S.~Shan, and X.~Chen, ``Single-side domain generalization for
  face anti-spoofing,'' in \emph{CVPR}, 2020.

\bibitem{akuzawa2019adversarial}
K.~Akuzawa, Y.~Iwasawa, and Y.~Matsuo, ``Adversarial invariant feature learning
  with accuracy constraint for domain generalization,'' in \emph{ECMLPKDD},
  2019.

\bibitem{aslani2020scanner}
S.~Aslani, V.~Murino, M.~Dayan, R.~Tam, D.~Sona, and G.~Hamarneh, ``Scanner
  invariant multiple sclerosis lesion segmentation from mri,'' in \emph{ISBI},
  2020.

\bibitem{zhao2020domain}
S.~Zhao, M.~Gong, T.~Liu, H.~Fu, and D.~Tao, ``Domain generalization via
  entropy regularization,'' in \emph{NeurIPS}, 2020.

\bibitem{li2020sequential}
D.~Li, Y.~Yang, Y.-Z. Song, and T.~Hospedales, ``Sequential learning for domain
  generalization,'' in \emph{ECCV-W}, 2020.

\bibitem{du2020learning}
Y.~Du, J.~Xu, H.~Xiong, Q.~Qiu, X.~Zhen, C.~G. Snoek, and L.~Shao, ``Learning
  to learn with variational information bottleneck for domain generalization,''
  in \emph{ECCV}, 2020.

\bibitem{du2021metanorm}
Y.~Du, X.~Zhen, L.~Shao, and C.~G.~M. Snoek, ``Metanorm: Learning to normalize
  few-shot batches across domains,'' in \emph{ICLR}, 2021.

\bibitem{wang2020meta}
B.~Wang, M.~Lapata, and I.~Titov, ``Meta-learning for domain generalization in
  semantic parsing,'' \emph{arXiv preprint arXiv:2010.11988}, 2020.

\bibitem{otalora2019staining}
S.~Ot{\'a}lora, M.~Atzori, V.~Andrearczyk, A.~Khan, and H.~M{\"u}ller,
  ``Staining invariant features for improving generalization of deep
  convolutional neural networks in computational pathology,'' \emph{Frontiers
  in bioengineering and biotechnology}, 2019.

\bibitem{chen2020improving}
C.~Chen, W.~Bai, R.~H. Davies, A.~N. Bhuva, C.~H. Manisty, J.~B. Augusto, J.~C.
  Moon, N.~Aung, A.~M. Lee, M.~M. Sanghvi \emph{et~al.}, ``Improving the
  generalizability of convolutional neural network-based segmentation on cmr
  images,'' \emph{Frontiers in cardiovascular medicine}, 2020.

\bibitem{zhang2020generalizing}
L.~Zhang, X.~Wang, D.~Yang, T.~Sanford, S.~Harmon, B.~Turkbey, B.~J. Wood,
  H.~Roth, A.~Myronenko, D.~Xu \emph{et~al.}, ``Generalizing deep learning for
  medical image segmentation to unseen domains via deep stacked
  transformation,'' \emph{TMI}, 2020.

\bibitem{qiao2020learning}
F.~Qiao, L.~Zhao, and X.~Peng, ``Learning to learn single domain
  generalization,'' in \emph{CVPR}, 2020.

\bibitem{sinha2017certifying}
A.~Sinha, H.~Namkoong, R.~Volpi, and J.~Duchi, ``Certifying some distributional
  robustness with principled adversarial training,'' \emph{arXiv preprint
  arXiv:1710.10571}, 2017.

\bibitem{xu2021robust}
Z.~Xu, D.~Liu, J.~Yang, C.~Raffel, and M.~Niethammer, ``Robust and
  generalizable visual representation learning via random convolutions,'' in
  \emph{ICLR}, 2021.

\bibitem{somavarapu2020frustratingly}
N.~Somavarapu, C.-Y. Ma, and Z.~Kira, ``Frustratingly simple domain
  generalization via image stylization,'' \emph{arXiv preprint
  arXiv:2006.11207}, 2020.

\bibitem{borlino2021rethinking}
F.~C. Borlino, A.~D'Innocente, and T.~Tommasi, ``Rethinking domain
  generalization baselines,'' \emph{arXiv preprint arXiv:2101.09060}, 2021.

\bibitem{carlucci2019hallucinating}
F.~M. Carlucci, P.~Russo, T.~Tommasi, and B.~Caputo, ``Hallucinating agnostic
  images to generalize across domains.'' in \emph{ICCV-W}, 2019.

\bibitem{xu2014exploiting}
Z.~Xu, W.~Li, L.~Niu, and D.~Xu, ``Exploiting low-rank structure from latent
  domains for domain generalization,'' in \emph{ECCV}, 2014.

\bibitem{niu2015multiview}
L.~Niu, W.~Li, and D.~Xu, ``Multi-view domain generalization for visual
  recognition,'' in \emph{ICCV}, 2015.

\bibitem{niu2015visual}
------, ``Visual recognition by learning from web data: A weakly supervised
  domain generalization approach,'' in \emph{CVPR}, 2015.

\bibitem{ding2017deep}
Z.~Ding and Y.~Fu, ``Deep domain generalization with structured low-rank
  constraint,'' \emph{TIP}, 2017.

\bibitem{d2018domain}
A.~D'Innocente and B.~Caputo, ``Domain generalization with domain-specific
  aggregation modules,'' in \emph{GCPR}, 2018.

\bibitem{mancini2018best}
M.~Mancini, S.~R. Bul{\`o}, B.~Caputo, and E.~Ricci, ``Best sources forward:
  domain generalization through source-specific nets,'' in \emph{ICIP}, 2018.

\bibitem{wang2020dofe}
S.~Wang, L.~Yu, K.~Li, X.~Yang, C.-W. Fu, and P.-A. Heng, ``Dofe:
  Domain-oriented feature embedding for generalizable fundus image segmentation
  on unseen datasets,'' \emph{TMI}, 2020.

\bibitem{seo2020learning}
S.~Seo, Y.~Suh, D.~Kim, J.~Han, and B.~Han, ``Learning to optimize domain
  specific normalization for domain generalization,'' in \emph{ECCV}, 2020.

\bibitem{segu2020batch}
M.~Seg{\`u}, A.~Tonioni, and F.~Tombari, ``Batch normalization embeddings for
  deep domain generalization,'' \emph{arXiv preprint arXiv:2011.12672}, 2020.

\bibitem{mancini2018robust}
M.~Mancini, S.~R. Bulo, B.~Caputo, and E.~Ricci, ``Robust place categorization
  with deep domain generalization,'' \emph{RA-L}, 2018.

\bibitem{cha2021domain}
J.~Cha, H.~Cho, K.~Lee, S.~Park, Y.~Lee, and S.~Park, ``Domain generalization
  needs stochastic weight averaging for robustness on domain shifts,''
  \emph{arXiv preprint arXiv:2102.08604}, 2021.

\bibitem{maniyar2020zero}
U.~Maniyar, A.~A. Deshmukh, U.~Dogan, and V.~N. Balasubramanian, ``Zero shot
  domain generalization,'' in \emph{BMVC}, 2020.

\bibitem{chattopadhyay2020learning}
P.~Chattopadhyay, Y.~Balaji, and J.~Hoffman, ``Learning to balance specificity
  and invariance for in and out of domain generalization,'' in \emph{ECCV},
  2020.

\bibitem{piratla2020efficient}
V.~Piratla, P.~Netrapalli, and S.~Sarawagi, ``Efficient domain generalization
  via common-specific low-rank decomposition,'' in \emph{ICML}, 2020.

\bibitem{ilse2019diva}
M.~Ilse, J.~M. Tomczak, C.~Louizos, and M.~Welling, ``Diva: Domain invariant
  variational autoencoder,'' in \emph{ICLR-W}, 2019.

\bibitem{wang2020cross}
G.~Wang, H.~Han, S.~Shan, and X.~Chen, ``Cross-domain face presentation attack
  detection via multi-domain disentangled representation learning,'' in
  \emph{CVPR}, 2020.

\bibitem{laskin2020reinforcement}
M.~Laskin, K.~Lee, A.~Stooke, L.~Pinto, P.~Abbeel, and A.~Srinivas,
  ``Reinforcement learning with augmented data,'' \emph{NeurIPS}, 2020.

\bibitem{kostrikov2020image}
I.~Kostrikov, D.~Yarats, and R.~Fergus, ``Image augmentation is all you need:
  Regularizing deep reinforcement learning from pixels,'' \emph{arXiv preprint
  arXiv:2004.13649}, 2020.

\bibitem{tobin2017domain}
J.~Tobin, R.~Fong, A.~Ray, J.~Schneider, W.~Zaremba, and P.~Abbeel, ``Domain
  randomization for transferring deep neural networks from simulation to the
  real world,'' in \emph{IROS}, 2017.

\bibitem{lee2019network}
K.~Lee, K.~Lee, J.~Shin, and H.~Lee, ``Network randomization: A simple
  technique for generalization in deep reinforcement learning,'' \emph{arXiv
  preprint arXiv:1910.05396}, 2019.

\bibitem{yarats2021improving}
D.~Yarats, A.~Zhang, I.~Kostrikov, B.~Amos, J.~Pineau, and R.~Fergus,
  ``Improving sample efficiency in model-free reinforcement learning from
  images,'' in \emph{AAAI}, 2021.

\bibitem{laskin2020curl}
M.~Laskin, A.~Srinivas, and P.~Abbeel, ``Curl: Contrastive unsupervised
  representations for reinforcement learning,'' in \emph{ICML}, 2020.

\bibitem{villani2008optimal}
C.~Villani, \emph{Optimal transport: old and new}.\hskip 1em plus 0.5em minus
  0.4em\relax Springer Science \& Business Media, 2008.

\bibitem{scholkopf2012causal}
B.~Sch{\"o}lkopf, D.~Janzing, J.~Peters, E.~Sgouritsa, K.~Zhang, and J.~Mooij,
  ``On causal and anticausal learning,'' in \emph{ICML}, 2012.

\bibitem{gretton2012kernel}
A.~Gretton, K.~M. Borgwardt, M.~J. Rasch, B.~Sch\"{o}lkopf, and A.~Smola, ``A
  kernel two-sample test,'' \emph{JMLR}, 2012.

\bibitem{goodfellow2014generative}
I.~Goodfellow, J.~Pouget-Abadie, M.~Mirza, B.~Xu, D.~Warde-Farley, S.~Ozair,
  A.~Courville, and Y.~Bengio, ``Generative adversarial nets,'' in
  \emph{NeurIPS}, 2014.

\bibitem{tzeng2017adversarial}
E.~Tzeng, J.~Hoffman, K.~Saenko, and T.~Darrell, ``Adversarial discriminative
  domain adaptation,'' in \emph{CVPR}, 2017.

\bibitem{zhang2018collaborative}
W.~Zhang, W.~Ouyang, W.~Li, and D.~Xu, ``Collaborative and adversarial network
  for unsupervised domain adaptation,'' in \emph{CVPR}, 2018.

\bibitem{long2018conditional}
M.~Long, Z.~Cao, J.~Wang, and M.~I. Jordan, ``Conditional adversarial domain
  adaptation,'' in \emph{NeurIPS}, 2018.

\bibitem{lee2019sliced}
C.-Y. Lee, T.~Batra, M.~H. Baig, and D.~Ulbricht, ``Sliced wasserstein
  discrepancy for unsupervised domain adaptation,'' in \emph{CVPR}, 2019.

\bibitem{ganin2016domain}
Y.~Ganin, E.~Ustinova, H.~Ajakan, P.~Germain, H.~Larochelle, F.~Laviolette,
  M.~Marchand, and V.~Lempitsky, ``Domain-adversarial training of neural
  networks,'' \emph{JMLR}, 2016.

\bibitem{finn2017model}
C.~Finn, P.~Abbeel, and S.~Levine, ``Model-agnostic meta-learning for fast
  adaptation of deep networks,'' in \emph{ICML}, 2017.

\bibitem{hospedales2020meta}
T.~Hospedales, A.~Antoniou, P.~Micaelli, and A.~Storkey, ``Meta-learning in
  neural networks: A survey,'' \emph{arXiv preprint arXiv:2004.05439}, 2020.

\bibitem{perez2020incremental}
J.-M. Perez-Rua, X.~Zhu, T.~M. Hospedales, and T.~Xiang, ``Incremental few-shot
  object detection,'' in \emph{CVPR}, 2020.

\bibitem{gordon2019meta}
J.~Gordon, J.~Bronskill, M.~Bauer, S.~Nowozin, and R.~E. Turner,
  ``Meta-learning probabilistic inference for prediction,'' in \emph{ICLR},
  2019.

\bibitem{goodfellow2016deep}
I.~Goodfellow, Y.~Bengio, and A.~Courville, \emph{Deep Learning}.\hskip 1em
  plus 0.5em minus 0.4em\relax MIT Press, 2016.

\bibitem{goodfellow2015explaining}
I.~J. Goodfellow, J.~Shlens, and C.~Szegedy, ``Explaining and harnessing
  adversarial examples,'' in \emph{ICLR}, 2015.

\bibitem{szegedy2014intriguing}
C.~Szegedy, W.~Zaremba, I.~Sutskever, J.~Bruna, D.~Erhan, I.~J. Goodfellow, and
  R.~Fergus, ``Intriguing properties of neural networks,'' in \emph{ICLR},
  2014.

\bibitem{huang2017arbitrary}
X.~Huang and S.~Belongie, ``Arbitrary style transfer in real-time with adaptive
  instance normalization,'' in \emph{ICCV}, 2017.

\bibitem{zhang2018mixup}
H.~Zhang, M.~Cisse, Y.~N. Dauphin, and D.~Lopez-Paz, ``mixup: Beyond empirical
  risk minimization,'' in \emph{ICLR}, 2018.

\bibitem{zhou2012ensemble}
Z.-H. Zhou, \emph{Ensemble methods: foundations and algorithms}.\hskip 1em plus
  0.5em minus 0.4em\relax Chapman and Hall/CRC, 2012.

\bibitem{szegedy2015going}
C.~Szegedy, W.~Liu, Y.~Jia, P.~Sermanet, S.~Reed, D.~Anguelov, D.~Erhan,
  V.~Vanhoucke, and A.~Rabinovich, ``Going deeper with convolutions,'' in
  \emph{CVPR}, 2015.

\bibitem{malisiewicz2011ensemble}
T.~Malisiewicz, A.~Gupta, and A.~A. Efros, ``Ensemble of exemplar-svms for
  object detection and beyond,'' in \emph{ICCV}, 2011.

\bibitem{ioffe2015batch}
S.~Ioffe and C.~Szegedy, ``Batch normalization: Accelerating deep network
  training by reducing internal covariate shift,'' in \emph{ICML}, 2015.

\bibitem{izmailov2018averaging}
P.~Izmailov, D.~Podoprikhin, T.~Garipov, D.~Vetrov, and A.~G. Wilson,
  ``Averaging weights leads to wider optima and better generalization,'' in
  \emph{UAI}, 2018.

\bibitem{jigsaw_puzzles}
M.~Noroozi and P.~Favaro, ``Unsupervised learning of visual representations by
  solving jigsaw puzzles,'' in \emph{ECCV}, 2016.

\bibitem{gidaris2018unsupervised}
S.~Gidaris, P.~Singh, and N.~Komodakis, ``Unsupervised representation learning
  by predicting image rotations,'' in \emph{ICLR}, 2018.

\bibitem{jing2020self}
L.~Jing and Y.~Tian, ``Self-supervised visual feature learning with deep neural
  networks: A survey,'' \emph{TPAMI}, 2020.

\bibitem{caron2018deep}
M.~Caron, P.~Bojanowski, A.~Joulin, and M.~Douze, ``Deep clustering for
  unsupervised learning of visual features,'' in \emph{ECCV}, 2018.

\bibitem{he2020momentum}
K.~He, H.~Fan, Y.~Wu, S.~Xie, and R.~Girshick, ``Momentum contrast for
  unsupervised visual representation learning,'' in \emph{CVPR}, 2020.

\bibitem{grill2020bootstrap}
J.-B. Grill, F.~Strub, F.~Altch{\'e}, C.~Tallec, P.~H. Richemond,
  E.~Buchatskaya, C.~Doersch, B.~A. Pires, Z.~D. Guo, M.~G. Azar \emph{et~al.},
  ``Bootstrap your own latent: A new approach to self-supervised learning,'' in
  \emph{NeurIPS}, 2020.

\bibitem{chen2016infogan}
X.~Chen, Y.~Duan, R.~Houthooft, J.~Schulman, I.~Sutskever, and P.~Abbeel,
  ``Infogan: Interpretable representation learning by information maximizing
  generative adversarial nets,'' in \emph{NeurIPS}, 2016.

\bibitem{devries2017cutout}
T.~DeVries and G.~W. Taylor, ``Improved regularization of convolutional neural
  networks with cutout,'' \emph{arXiv preprint arXiv:1708.04552}, 2017.

\bibitem{redko2020survey}
I.~Redko, E.~Morvant, A.~Habrard, M.~Sebban, and Y.~Bennani, ``A survey on
  domain adaptation theory: learning bounds and theoretical guarantees,''
  \emph{arXiv preprint arXiv:2004.11829}, 2020.

\bibitem{deshmukh2019generalization}
A.~A. Deshmukh, Y.~Lei, S.~Sharma, U.~Dogan, J.~W. Cutler, and C.~Scott, ``A
  generalization error bound for multi-class domain generalization,''
  \emph{arXiv preprint arXiv:1905.10392}, 2019.

\bibitem{rosenfeld2022online}
E.~Rosenfeld, P.~Ravikumar, and A.~Risteski, ``An online learning approach to
  interpolation and extrapolation in domain generalization,'' in
  \emph{AISTATS}, 2022.

\bibitem{vedantam2021empirical}
R.~Vedantam, D.~Lopez-Paz, and D.~J. Schwab, ``An empirical investigation of
  domain generalization with empirical risk minimizers,'' \emph{NeurIPS}, 2021.

\bibitem{ye2021towards}
H.~Ye, C.~Xie, T.~Cai, R.~Li, Z.~Li, and L.~Wang, ``Towards a theoretical
  framework of out-of-distribution generalization,'' \emph{NeurIPS}, 2021.

\bibitem{li2022finding}
D.~Li, H.~Gouk, and T.~Hospedales, ``Finding lost dg: Explaining domain
  generalization via model complexity,'' \emph{arXiv preprint
  arXiv:2202.00563}, 2022.

\bibitem{han2021dynamic}
Y.~Han, G.~Huang, S.~Song, L.~Yang, H.~Wang, and Y.~Wang, ``Dynamic neural
  networks: A survey,'' \emph{arXiv preprint arXiv:2102.04906}, 2021.

\bibitem{jia2016dynamic}
X.~Jia, B.~De~Brabandere, T.~Tuytelaars, and L.~Van~Gool, ``Dynamic filter
  networks,'' in \emph{NeurIPS}, 2016.

\bibitem{yang2019condconv}
B.~Yang, G.~Bender, Q.~V. Le, and J.~Ngiam, ``Condconv: Conditionally
  parameterized convolutions for efficient inference,'' in \emph{NeurIPS},
  2019.

\bibitem{ulyanov2016instance}
D.~Ulyanov, A.~Vedaldi, and V.~Lempitsky, ``Instance normalization: The missing
  ingredient for fast stylization,'' \emph{arXiv:1607.08022}, 2016.

\bibitem{ba2016layer}
J.~L. Ba, J.~R. Kiros, and G.~E. Hinton, ``Layer normalization,'' \emph{arXiv
  preprint arXiv:1607.06450}, 2016.

\bibitem{luo2019differentiable}
P.~Luo, J.~Ren, Z.~Peng, R.~Zhang, and J.~Li, ``Differentiable
  learning-to-normalize via switchable normalization,'' in \emph{ICLR}, 2019.

\bibitem{park2019semantic}
T.~Park, M.-Y. Liu, T.-C. Wang, and J.-Y. Zhu, ``Semantic image synthesis with
  spatially-adaptive normalization,'' in \emph{CVPR}, 2019.

\bibitem{deecke2020latent}
L.~Deecke, T.~Hospedales, and H.~Bilen, ``Latent domain learning with dynamic
  residual adapters,'' \emph{arXiv preprint arXiv:2006.00996}, 2020.

\bibitem{xu2021how}
K.~Xu, M.~Zhang, J.~Li, S.~S. Du, K.-I. Kawarabayashi, and S.~Jegelka, ``How
  neural networks extrapolate: From feedforward to graph neural networks,'' in
  \emph{ICLR}, 2021.

\bibitem{geirhos2020shortcut}
R.~Geirhos, J.-H. Jacobsen, C.~Michaelis, R.~Zemel, W.~Brendel, M.~Bethge, and
  F.~A. Wichmann, ``Shortcut learning in deep neural networks,'' \emph{Nature
  Machine Intelligence}, 2020.

\bibitem{kim2019learning}
B.~Kim, H.~Kim, K.~Kim, S.~Kim, and J.~Kim, ``Learning not to learn: Training
  deep neural networks with biased data,'' in \emph{CVPR}, 2019.

\bibitem{bengio2019meta}
Y.~Bengio, T.~Deleu, N.~Rahaman, R.~Ke, S.~Lachapelle, O.~Bilaniuk, A.~Goyal,
  and C.~Pal, ``A meta-transfer objective for learning to disentangle causal
  mechanisms,'' \emph{arXiv preprint arXiv:1901.10912}, 2019.

\bibitem{scholkopf2021towards}
B.~Scholkopf, F.~Locatello, S.~Bauer, N.~R. Ke, N.~Kalchbrenner, A.~Goyal, and
  Y.~Bengio, ``Towards causal representation learning,'' \emph{arXiv preprint
  arXiv:2102.11107}, 2021.

\bibitem{hoffman2016learning}
J.~Hoffman, S.~Gupta, and T.~Darrell, ``Learning with side information through
  modality hallucination,'' in \emph{CVPR}, 2016.

\bibitem{zhou2017detecting}
K.~Zhou, A.~Paiement, and M.~Mirmehdi, ``Detecting humans in rgb-d data with
  cnns,'' in \emph{MVA}, 2017.

\bibitem{zunino2020explainable}
A.~Zunino, S.~A. Bargal, R.~Volpi, M.~Sameki, J.~Zhang, S.~Sclaroff, V.~Murino,
  and K.~Saenko, ``Explainable deep classification models for domain
  generalization,'' \emph{arXiv preprint arXiv:2003.06498}, 2020.

\bibitem{zeiler2014visualizing}
M.~D. Zeiler and R.~Fergus, ``Visualizing and understanding convolutional
  networks,'' in \emph{ECCV}, 2014.

\bibitem{li2017learning}
Z.~Li and D.~Hoiem, ``Learning without forgetting,'' \emph{TPAMI}, 2017.

\bibitem{sharifi2020domain}
H.~Sharifi-Noghabi, H.~Asghari, N.~Mehrasa, and M.~Ester, ``Domain
  generalization via semi-supervised meta learning,'' \emph{arXiv preprint
  arXiv:2009.12658}, 2020.

\bibitem{liu2021semi}
X.~Liu, S.~Thermos, A.~O’Neil, and S.~A. Tsaftaris, ``Semi-supervised
  meta-learning with disentanglement for domain-generalised medical image
  segmentation,'' in \emph{MICCAI}, 2021.

\bibitem{liu2022vmfnet}
X.~Liu, S.~Thermos, P.~Sanchez, A.~Q. O'Neil, and S.~A. Tsaftaris, ``vmfnet:
  Compositionality meets domain-generalised segmentation,'' in \emph{MICCAI},
  2022.

\bibitem{geng2020recent}
C.~Geng, S.-j. Huang, and S.~Chen, ``Recent advances in open set recognition: A
  survey,'' \emph{TPAMI}, 2020.

\bibitem{wu2019large}
Y.~Wu, Y.~Chen, L.~Wang, Y.~Ye, Z.~Liu, Y.~Guo, and Y.~Fu, ``Large scale
  incremental learning,'' in \emph{CVPR}, 2019.

\bibitem{li2014deepreid}
W.~Li, R.~Zhao, T.~Xiao, and X.~Wang, ``Deepreid: Deep filter pairing neural
  network for person re-identification,'' in \emph{CVPR}, 2014.

\end{thebibliography}
}

%



%

\begin{IEEEbiography}[{\includegraphics[width=1in,height=1.25in,clip,keepaspectratio]{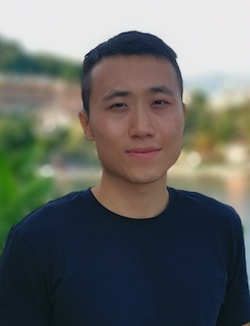}}]{Kaiyang Zhou}
received the PhD degree in Computer Science from the University of Surrey, UK, in 2020. He is currently a Research Fellow at Nanyang Technological University, Singapore. His research lies at the intersection of machine learning and computer vision. His papers have been published at major journals and conferences in relevant fields, such as TPAMI, IJCV, ICLR, AAAI, CVPR, ICCV, and ECCV. According to Google Scholar, his papers have been cited more than 1,600 times, with h-index at 14. He serves/served as an Area Chair/Senior Program Committee Member for BMVC (2022) and AAAI (2023), and a reviewer for top-tier journals and conferences including TPAMI, ICLR, NeurIPS, ICML, CVPR, ICCV, ECCV, etc.
\end{IEEEbiography}

\begin{IEEEbiography}[{\includegraphics[width=1in,height=1.25in,clip,keepaspectratio]{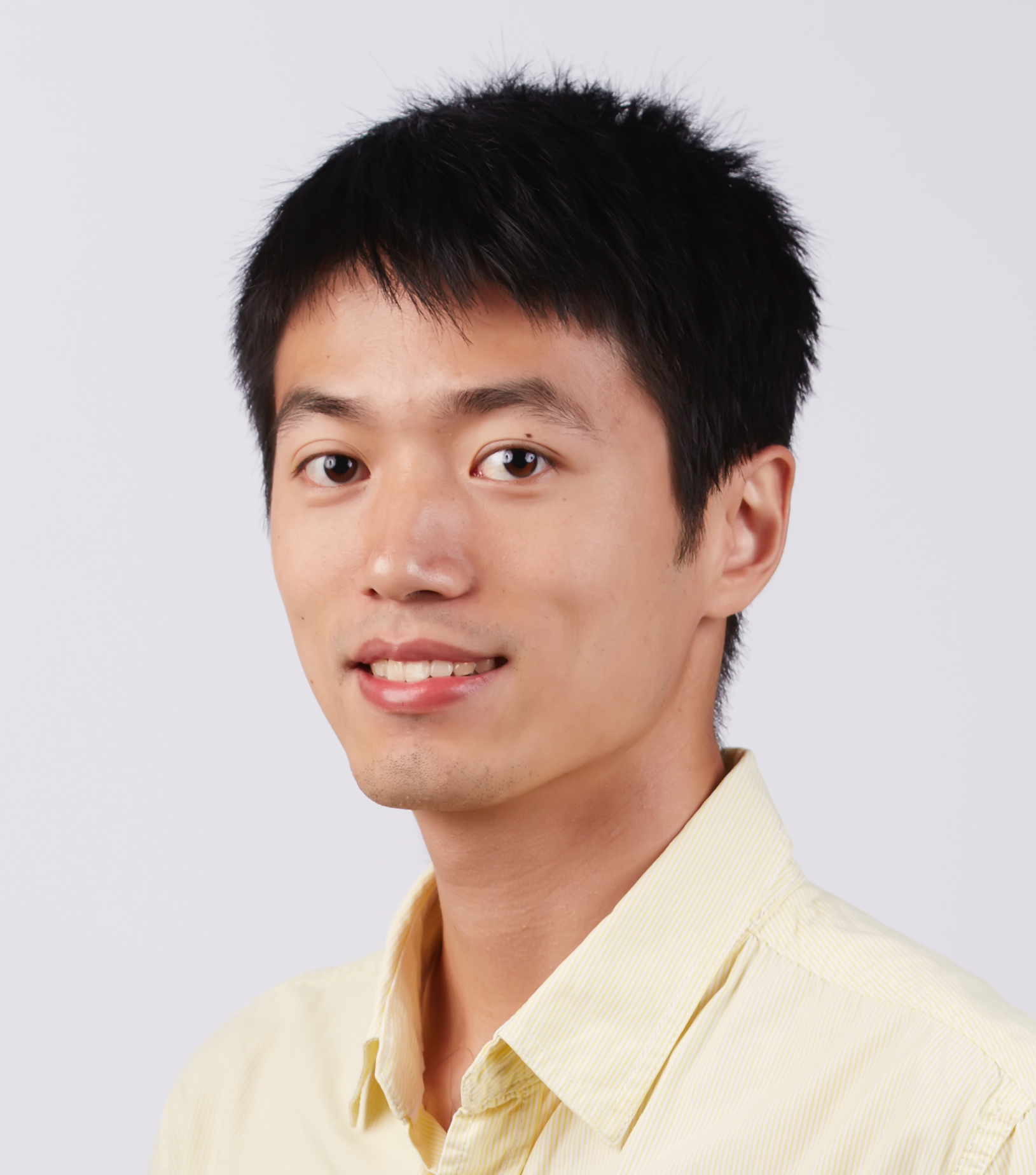}}]{Ziwei Liu}
is currently an Assistant Professor at Nanyang Technological University, Singapore. Previously, he was a senior research fellow at the Chinese University of Hong Kong and a postdoctoral researcher at University of California, Berkeley. Ziwei received his PhD from the Chinese University of Hong Kong. His research revolves around computer vision, machine learning and computer graphics. He has published extensively on top-tier conferences and journals in relevant fields, including CVPR, ICCV, ECCV, NeurIPS, ICLR, ICML, TPAMI, TOG and Nature - Machine Intelligence. He is the recipient of Microsoft Young Fellowship, Hong Kong PhD Fellowship, ICCV Young Researcher Award and HKSTP Best Paper Award. He also serves as an Area Chair of ICCV, NeurIPS and AAAI.
\end{IEEEbiography}

\begin{IEEEbiography}[{\includegraphics[width=1in,height=1.25in,clip,keepaspectratio]{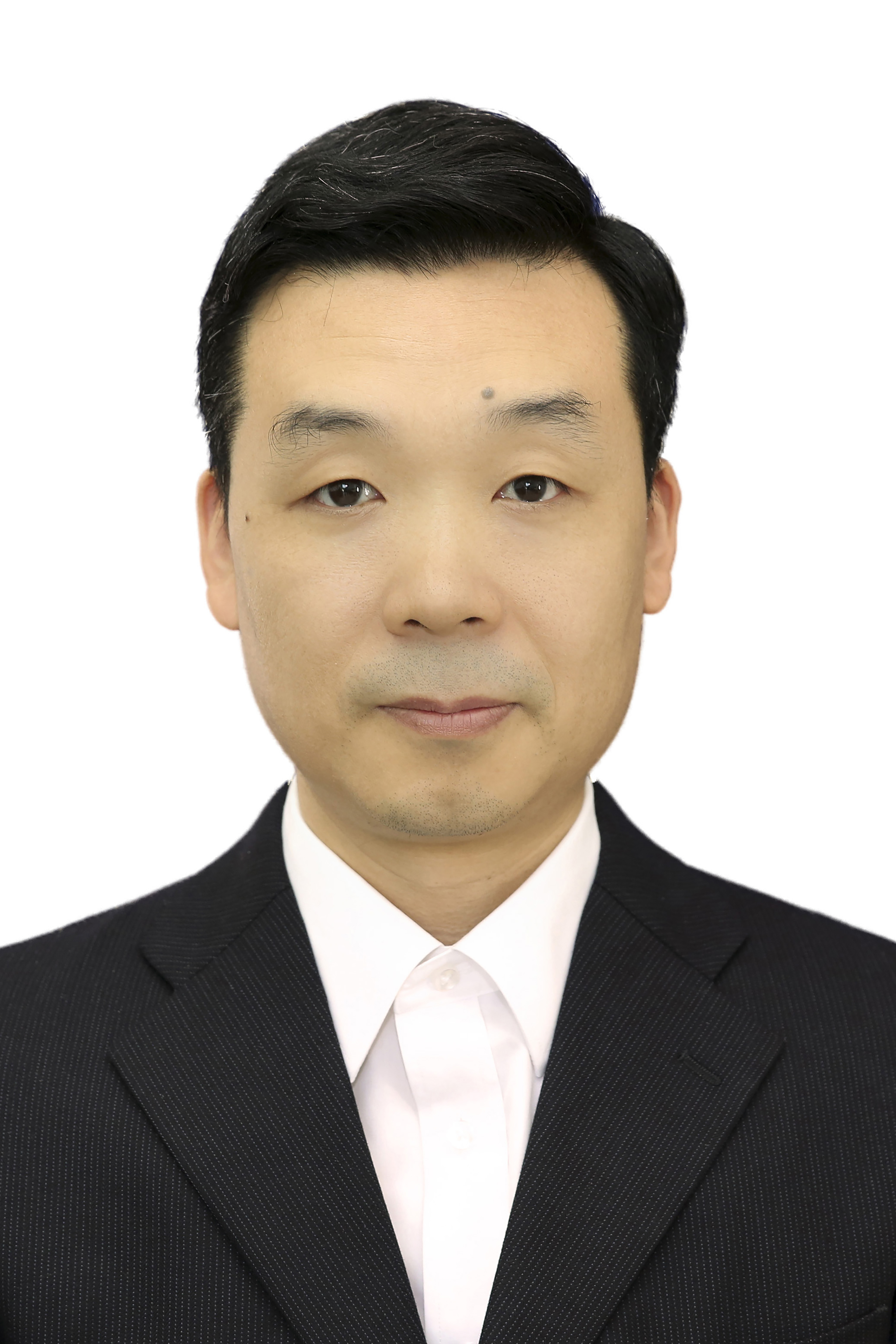}}]{Yu Qiao}
is a professor with Shanghai AI Laboratory and the Shenzhen Institutes of Advanced Technology (SIAT), the Chinese Academy of Sciences. His research interests include computer vision, deep learning, and bioinformation. He has published more than 300 papers in international journals and conferences, including T-PAMI, IJCV, T-IP, T-SP, CVPR, ICCV, etc. His H-index is 72, with 35,000 citations in Google scholar. He is the recipient of the distinguished paper award in AAAI 2021. His group achieved the first runner-up at the ImageNet Large Scale Visual Recognition Challenge 2015 in scene recognition, and was the winner at the ActivityNet Large Scale Activity Recognition Challenge 2016 in video classification. He served as the program chair of IEEE ICIST 2014.
\end{IEEEbiography}

\begin{IEEEbiography}[{\includegraphics[width=1in,height=1.25in,clip,keepaspectratio]{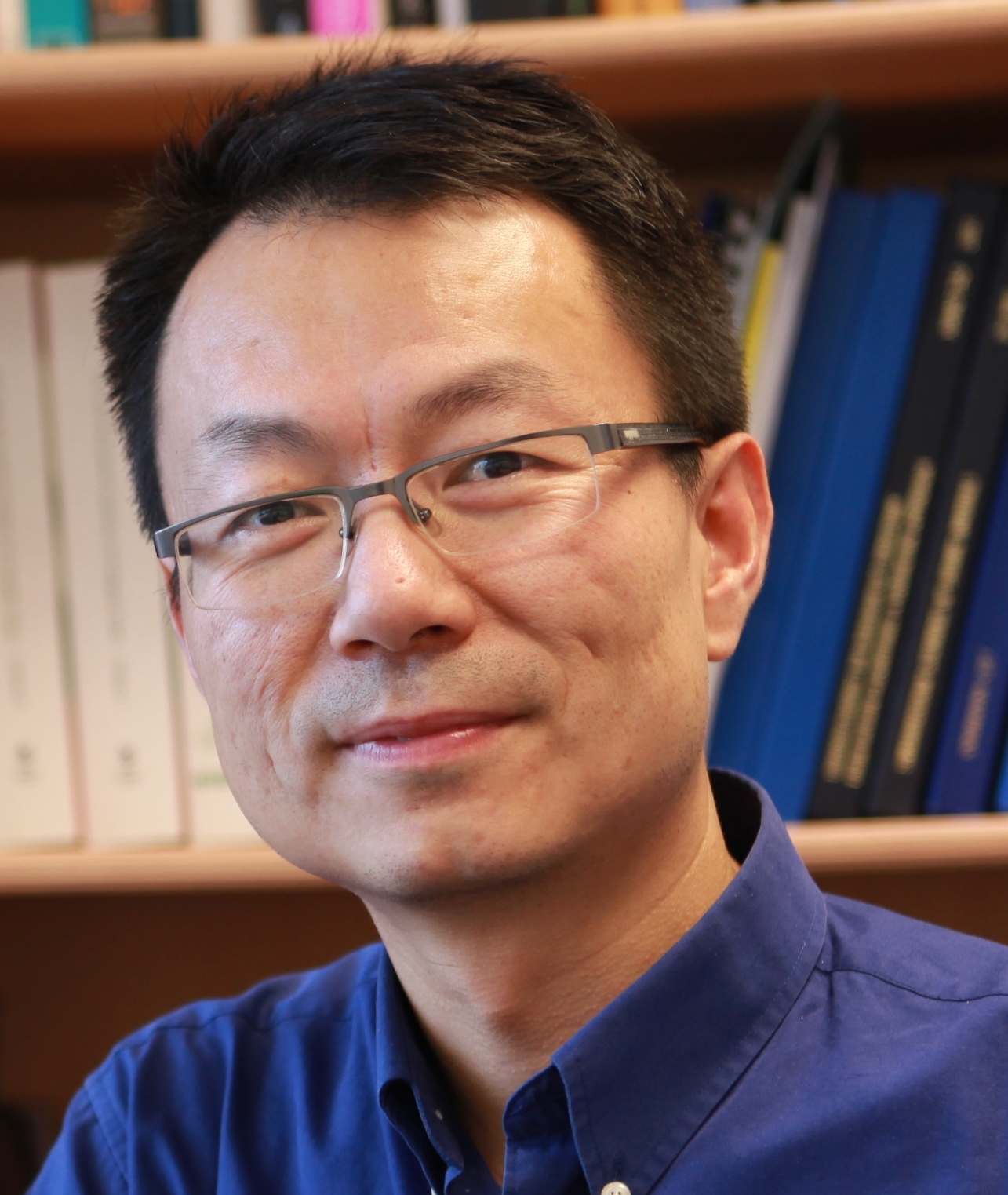}}]{Tao Xiang}
received the Ph.D. degree in electrical and computer engineering from the National University of Singapore in 2002. He is currently a full professor in the Department of Electrical and Electronic Engineering, University of Surrey and a Research Scientist Manager at Meta AI. His research interests include computer vision and machine learning. He has published over 200 papers in international journals and conferences with over 28K citations.
\end{IEEEbiography}

\begin{IEEEbiography}[{\includegraphics[width=1in,height=1.25in,clip,keepaspectratio]{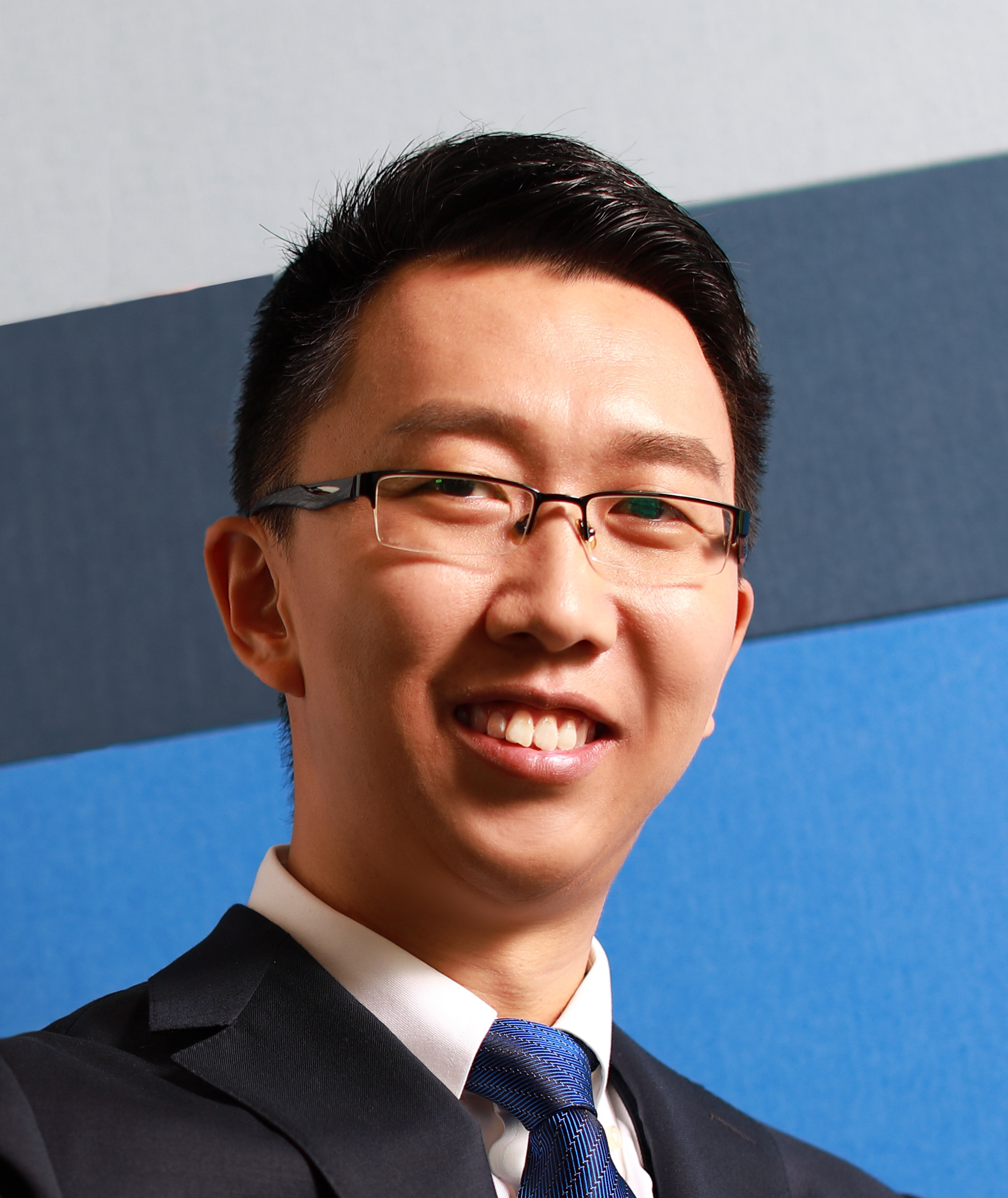}}]{Chen Change Loy} (Senior Member, IEEE) received the PhD degree in computer science from the Queen Mary University of London, in 2010. He is an associate professor with the School of Computer Science and Engineering, Nanyang Technological University. Prior to joining NTU, he served as a research assistant professor with the Department of Information Engineering, The Chinese University of Hong Kong, from 2013 to 2018. His research interests include computer vision and deep learning with a focus on image/video restoration and enhancement, generative tasks, and representation learning. He serves as an Associate Editor of the International Journal of Computer Vision (IJCV) and IEEE Transactions on Pattern Analysis and Machine Intelligence (TPAMI). He also serves/served as an Area Chair of ICCV 2021, CVPR (2021, 2019), ECCV (2022, 2018), AAAI (2021-2023), and BMVC (2018-2021).
\end{IEEEbiography}




\end{document}